\documentclass[11pt]{article}

\usepackage[preprint]{acl}

\usepackage{times}
\usepackage{latexsym}

\usepackage[T1]{fontenc}

\usepackage[utf8]{inputenc}

\usepackage{microtype}
\usepackage{booktabs} 
\usepackage{array}
\usepackage{needspace}
\usepackage{float}

\makeatletter
\g@addto@macro\UrlBreaks{%
  \do\a\do\b\do\c\do\d\do\e\do\f\do\g\do\h\do\i\do\j\do\k\do\l\do\m%
  \do\n\do\o\do\p\do\q\do\r\do\s\do\t\do\u\do\v\do\w\do\x\do\y\do\z%
  \do\A\do\B\do\C\do\D\do\E\do\F\do\G\do\H\do\I\do\J\do\K\do\L\do\M%
  \do\N\do\O\do\P\do\Q\do\R\do\S\do\T\do\U\do\V\do\W\do\X\do\Y\do\Z}
\makeatother
\usepackage{amssymb}
\usepackage{inconsolata}

\usepackage{graphicx}
\usepackage{multirow}
\usepackage{amsmath}
\usepackage{listings}
\usepackage{subfiles}
\usepackage{pgfplots}
\usepackage{tikz}
\usetikzlibrary{shapes, shapes.geometric, positioning, fit, calc, arrows.meta, backgrounds, matrix, decorations.pathreplacing}
\usepackage{graphicx} 
\pgfplotsset{compat=1.18}
\usepackage{booktabs}
\usepackage{multirow} 
\usepackage{geometry}
 
\usepackage{amsmath}
\definecolor{plotblue}{RGB}{31,119,180}
\newcommand{\datasetname}{\textsc{NORA Dataset}}

\definecolor{posgreen}{RGB}{0,140,0}
\definecolor{negred}{RGB}{200,0,0}
\newcommand{\dplus}[1]{\textcolor{posgreen}{\scriptsize(+#1)}}

\title{Noise-Robust Financial Numerical Entity Attribute Tagging}

\author{
  \textbf{Sheng-Hua Tsai\textsuperscript{1}},
  \textbf{Chen-Yang Lai\textsuperscript{1}},
  \textbf{Yi-Jhen Li\textsuperscript{1}},
\\
  \textbf{Ju-Chun Yen\textsuperscript{2}},
  \textbf{Hsin-Min Lu\textsuperscript{1}}
\\
  \textsuperscript{1}National Taiwan University,
  \textsuperscript{2}National Central University
\\
  \small{\texttt{kmes80214@gmail.com}, \texttt{r13725064@ntu.edu.tw}, \texttt{yjli725049@gmail.com},}
\\
  \small{\texttt{jcyen@ncu.edu.tw}, \texttt{luim@ntu.edu.tw}}
}

\begin{document}
\maketitle
\begin{abstract}
Financial Numerical Entity (FNE) understanding aims to recover the meaning of numerical mentions in financial reports. Existing studies focus on concept name prediction and face two important limitations. First, labels derived from inline XBRL may contain errors because filings are usually prepared manually. Second, other important FNE attributes, such as reporting-time relation, measurement scale, and accounting sign, are less emphasized.
We propose \textbf{NO}ise-\textbf{R}obust Tagging for Rich Financial Numerical Entity \textbf{A}ttributes (\textsc{NORA}) to address these gaps. NORA uses task-aware instance-specific weighting to attenuate the influence of noisy labels during training, and we develop an evaluation protocol that filters noisy labels in real-world test sets. In addition, we construct a large-scale benchmark containing 11.4 million FNEs in 6.3 million sentences with multi-attribute labels and filing metadata.
Experiments show that \textsc{NORA} performs strongly compared with previous financial numeric tagging models and noisy-label baselines. Under both unfiltered and noise-filtered test settings, it achieves the best mean Accuracy and Macro F1 over the four attributes, with the largest gain on time-relation prediction, while remaining competitive on concept name, scale, and sign. These results demonstrate the value of modeling rich FNE attributes with document context while accounting for label noise in real-world financial filings.
\end{abstract}

\section{Introduction}
\label{sec:introduction}

\begin{figure}[htbp]
    \centering
    \includegraphics[width=0.47\textwidth]{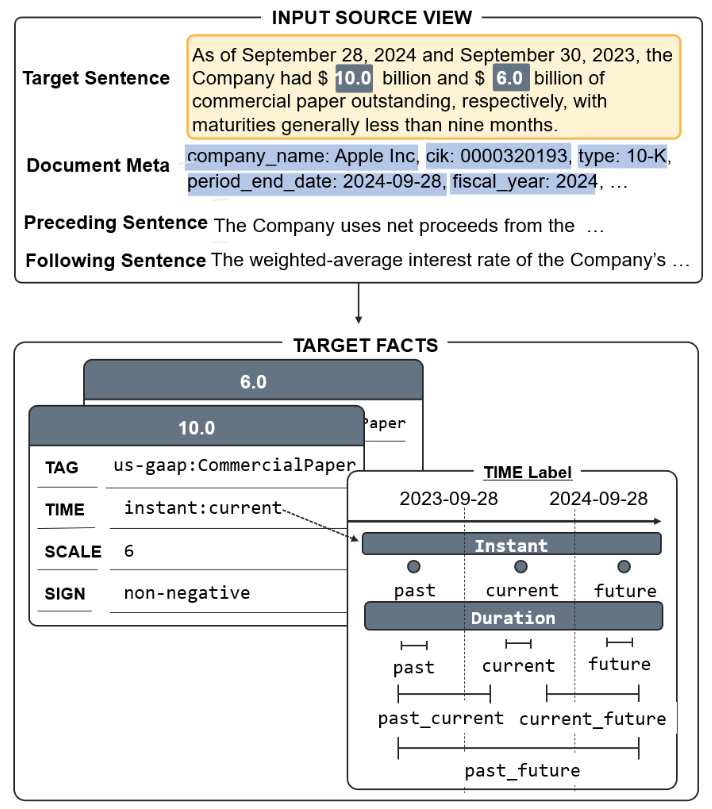}
    \caption{Illustration of input context and corresponding
FNE attributes.}
    \label{fig:input_target}
\end{figure}

Financial Numerical Entity (FNE) understanding is the task of identifying the meaning of numerical mentions in financial text by predicting the attributes that make them interpretable in context \citep{huang2023finbert, yang2020finbert}. In financial reports such as 10-Ks or 10-Qs, this means determining not only the accounting concept (i.e., the tag) an FNE is associated with, but also attributes such as whether it is an instant or duration fact, its relation to the current reporting period (e.g., past, current, or future), its measurement scale (e.g., million), and the sign of the value with respect to the accounting concept (see Figure \ref{fig:input_target} for an illustrative example). The problem is important because downstream systems for risk analysis, compliance monitoring, and business analysis all depend on assigning the correct semantics to numbers in disclosure documents \citep{fourny2023xbrl}.

Existing studies have laid important groundwork for FNE understanding. Benchmarks such as FiNER-139 and FNXL have advanced financial entity tagging by standardizing datasets and evaluations for identifying numerical entities and linking them to accounting concepts \citep{loukas2022finer, sharma2023fnxl}. However, two important gaps remain. First, large-scale supervision in this domain is often obtained from inline XBRL (iXBRL), yet iXBRL is inherently noisy as a source for ground truth. Several studies document recurring filing mistakes such as mismatches between embedded tags and surrounding narrative, missing elements, incorrect monetary values, and sign errors \citep{bartley2011comparison, du2013xbrl}. Second, prior benchmarks largely reduce the task to concept-name tagging and do not jointly model other crucial attributes such as time relation, measurement scale, and the accounting sign of numeric entities. As a result, existing formulations understate both the semantic richness of the task and the practical difficulty of learning from noisy labels.

To address these limitations, we propose the \textbf{NO}ise-\textbf{R}obust Tagging for Rich Financial Numerical Entity \textbf{A}ttributes (NORA) framework. NORA treats FNE understanding as a joint prediction problem over related attributes, including concept name (i.e., tag), reporting-time relation, measurement scale, and accounting sign, and is designed to be robust to noisy labels induced by real-world iXBRL filings. The key idea is to incorporate learnable instance-specific weighting so that the model can attenuate the overfitting downside of noisy labels while still benefiting from large-scale automatically derived supervision. NORA also draws on the filing metadata and adjacent sentences provided by our dataset, which supply context beyond the target sentence. For evaluation, we develop a protocol that filters noisy labels in real-world test sets.
As a companion dataset to NORA, we curate a large-scale corpus of 6.3 million narrative instances aligned to iXBRL facts with rich FNE attributes from nearly 110{,}000 10-K and 10-Q filings, directly targeting the gap between idealized benchmark settings and realistic financial reporting environments.

Our main contributions are as follows:
\begin{itemize}
    \item We formulate FNE understanding as a rich multi-attribute prediction task that jointly models concept name, reporting-time relation, measurement scale, and accounting sign.
    \item We construct a large-scale benchmark from real-world iXBRL filings, comprising 6.3 million narrative instances with rich FNE attributes aligned to facts from nearly 110{,}000 SEC filings.
    \item We propose NORA, a noise-robust learning framework that uses learnable instance-specific weighting to mitigate overfitting to noisy supervision and consistently outperforms strong noisy-label learning baselines on most tasks.
    \item We develop an evaluation protocol that filters noisy test labels and enables more reliable model evaluation using real-world noisy filings.
\end{itemize}

\section{Related Works}
\label{sec:related_work}

\subsection{Financial Numerical Entity (FNE) Tagging} 

FNEs are numerical mentions in financial reports whose meanings are determined by the surrounding text \citep{loukas2022finer, sharma2023fnxl}. Approximately 91\% of tagged tokens in these reports are numeric, meaning the correct label depends almost entirely on the surrounding textual and structured reporting attributes \citep{loukas2022finer}. How language models represent numbers and their magnitudes also remains an open problem \citep{thawani-etal-2021-representing, zhang-etal-2020-language-embeddings}, which bears directly on scale prediction.

FiNER-139 standardized evaluation for the 139 most frequent XBRL tags \citep{loukas2022finer}, and the Financial Numeric Extreme Labelling (FNXL) dataset expands the label set to 2,794 distinct tags \citep{sharma2023fnxl}. Together, these resources have established FNE tagging as an extreme classification problem, characterized by a heavy-tail label distribution where many tags appear very infrequently. 
The FLAN-FinXC \citep{khatuya2024parameter} took a generative approach by instruction-tuned a FLAN-T5 to output XBRL tag documentations given a FNE in context. An unsupervised tag matcher then output the final concept name. 

However, most formulations emphasize the span and/or concept name of a numerical mention, while the complete interpretation of FNEs in filings also depends on attributes such as reporting time, measurement scale, and accounting sign.
A second limitation is that filing supervision is naturally noisy. Prior work documents quality issues in XBRL reporting, including reporting inconsistencies, missing elements, incorrect monetary values, and sign mistakes \citep{bartley2011comparison, du2013xbrl}. Filings that use more XBRL tags are also more likely to contain misstatements \citep{10.2308/accr-51762}. These concerns remain relevant in iXBRL settings, where the same concept can be tagged differently across firms \citep{chang2021effect}. Recent studies further show that the adoption of XBRL and iXBRL affects information transparency and market outcomes \citep{alokaily2024xbrl, luo2023initial}. Because large-scale FNE labels are derived from iXBRL facts, training and evaluation on noisy ground truth may lead to sub-optimal or even misleading results.
 In contrast to prior work, we formulate FNE tagging as joint attribute prediction over numerical mentions and study it under naturally noisy, document-grounded supervision.

\subsection{Robust Learning under Noisy Labels}
Learning from noisy supervision has been studied extensively in deep learning and NLP, but its implications for FNE tagging remain underexplored. Representative approaches include \textit{mixup}, which regularizes training through interpolation \citep{zhang2017mixup}. Co-teaching exchanges small-loss examples between two networks \citep{han2018coteaching}. SelfMix combines sample partitioning with semi-supervised learning for text classification \citep{qiao2022selfmix}. SSR uses neighbor-based sample selection and subset expansion for efficient noisy-label learning \citep{Feng_2022_BMVC, 10594806}. A broader survey is provided by \citet{9729424}. More recent studies leverage large language models (LLMs) to identify or correct noisy labels \citep{wang-etal-2023-noise, yuan-etal-2024-hide}. We do not adopt this approach because querying an LLM for millions of instances is costly, a limitation that \citet{wang-etal-2023-noise} also acknowledge.

Among these directions, loss reweighting is particularly relevant to FNE tagging because it directly controls how much each training instance contributes to optimization. Prior importance-reweighting work shows that, under symmetric random classification noise and mild assumptions, a properly reweighted loss can recover the classifier that would have been learned from clean data \citep{7929355}. However, such approaches often rely on fixed reweighting functions or pre-specified noise rates, which are difficult to justify for iXBRL-derived supervision.

In financial filings, label reliability is unlikely to be uniform across either instances or attributes. A numerical mention may therefore be reliable for one attribute, such as time relations, but mislabeled for another, such as concept names. This setting calls for instance-specific and task-aware loss reweighting, yet this direction has received limited attention in noisy label studies in general and in FNE tagging in particular. Our work addresses this gap by using learnable instance-specific weights to reduce the influence of noisy labels during multi-attribute FNE tagging.

Noisy supervision also creates a practical evaluation challenge. Most noisy-label studies assume access to a clean test set \citep{9729424}, but this assumption is often unrealistic for large-scale financial filings, where manually constructing a noise-free evaluation set is expensive and difficult to scale. If test labels are derived from the same imperfect iXBRL sources as training labels, benchmark scores may reflect both model performance and  annotation noise. To mitigate this issue, we develop an evaluation protocol, inspired by neighbor-based sample selection \citep{Feng_2022_BMVC, 10594806}, that constructs more reliable evaluation subsets from real-world noisy filings.

\section{Methodology}
\label{sec:methodology}

We first present noise-robust training in NORA and then the evaluation protocol for noisy test labels.

\subsection{Noise-Robust Training in NORA}

\begin{figure*}[t]
\centering
\includegraphics[width=\textwidth]{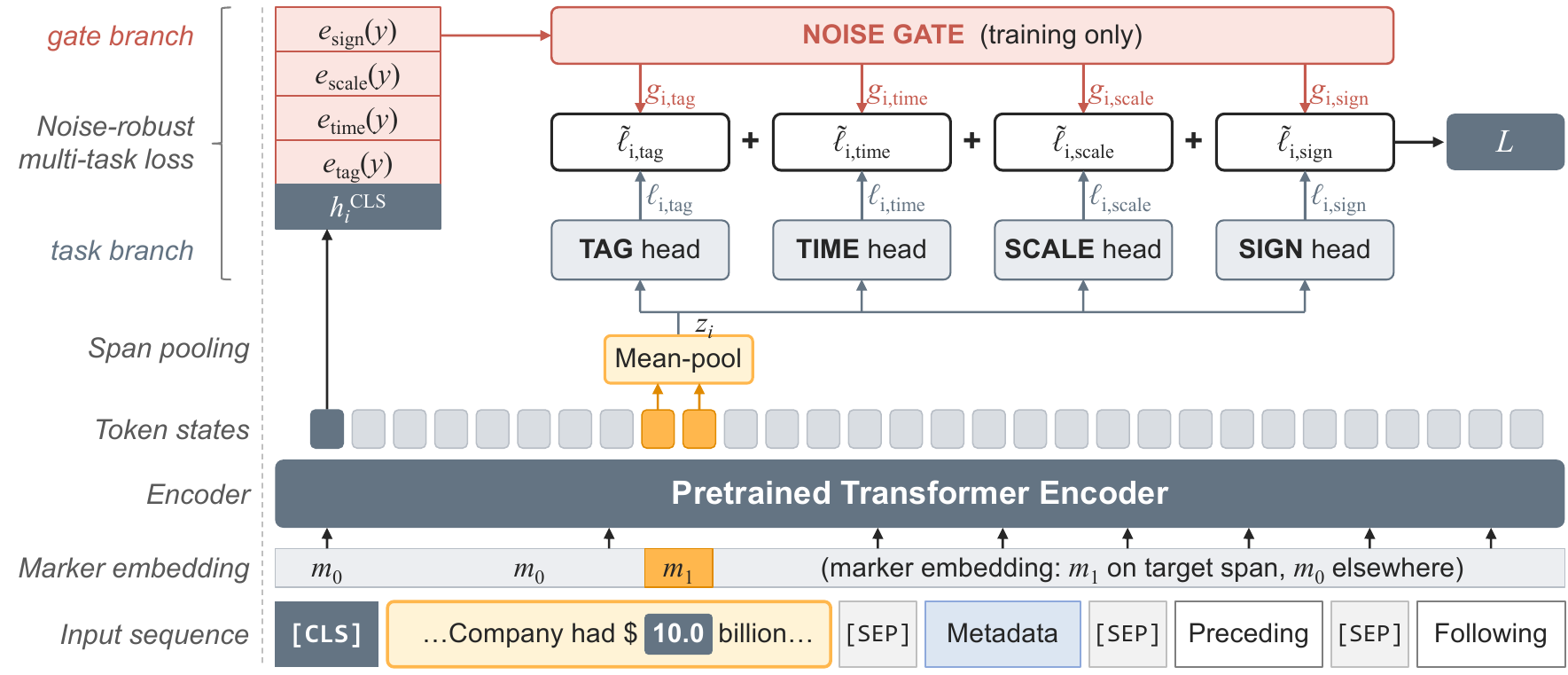}
\caption{Overview of NORA. The input concatenates the target sentence, filing metadata, and the preceding and following sentences, and a marker embedding identifies the target span. Each task head reads the mean-pooled span representation $\mathbf{z}_i$. During training, the noise gate reads the \texttt{[CLS]} representation and the label embeddings and outputs a gate value $g_{i,a}$ for each task, which reweights the task loss $\ell_{i,a}$. The gate is not used at inference.}
\label{fig:model}
\end{figure*}

NORA builds on a pretrained transformer encoder \citep{devlin-etal-2019-bert} and adds a reweighting component to each classification head to reduce the downside of training with noisy labels (Figure~\ref{fig:model}). The input sequence consists of \texttt{[CLS]}, the target sentence, filing metadata, the preceding sentence, and the following sentence, separated by \texttt{[SEP]}. The metadata includes the company name, CIK, document type (e.g., 10-K or 10-Q), report-period end date, and current fiscal year. If the resulting sequence exceeds the context window, we truncate it to fit the encoder limit.

We treat the location of the FNE span as given to isolate the effect of noise-robust attribute prediction.

To make the encoder aware of which numeral is the target, NORA adds to the input token embeddings a learned marker embedding that assigns $\mathbf{m}_1$ to tokens in the target span and $\mathbf{m}_0$ to all other tokens. Both vectors are initialized to zero. Without the marker, FNEs in the same sentence share an identical input sequence and therefore an identical \texttt{[CLS]} representation.

Because label reliability may depend on the full textual and metadata context and on the labels themselves, NORA estimates an instance-specific noise level from the \texttt{[CLS]} representation together with embeddings of the instance's labels. Let $\mathbf{h}_i^{\mathrm{CLS}}$ denote the contextual representation of the \texttt{[CLS]} token of instance $i$ and $\mathbf{e}_a(y_{i,a})$ a learned embedding of its label for task $a$, with one additional embedding for a missing label. We concatenate them, project the result through a shared dense layer with $\tanh$ activation, and then apply a task-specific sigmoid layer. For task $a$, the estimated noise gate of instance $i$ is:
\begin{equation*}
\begin{aligned}
    \mathbf{u}_i &= [\mathbf{h}_i^{\mathrm{CLS}}; \mathbf{e}_{\texttt{tag}}(y_{i,\texttt{tag}}); \dots; \mathbf{e}_{\texttt{sign}}(y_{i,\texttt{sign}})], \\
    \tilde{\mathbf{h}}_i &= \tanh(\mathbf{W}\mathbf{u}_i + \mathbf{b}), \\
    g_{i,a} &= \sigma(\mathbf{W}_a\tilde{\mathbf{h}}_i + b_a).
\end{aligned}
\end{equation*}
We intend to integrate these label embeddings with the sentences to urge the model to discern which combinations of labels are valid and which may be dubious. Without this, a combination of correct labels and a combination of noisy labels may result in the same gate value given the same instance due to lack of label information. The label embeddings are initialized to zero and enter only the gate, not the task heads. Because the gate only weights the training loss, labels are neither required nor leaked at inference.

\noindent Since $g_{i,a} \in (0,1)$, it can be interpreted as the estimated noise level of instance $i$ for task $a$. When $g_{i,a} \approx 0$, the instance is treated as reliable and informative for that task. When $g_{i,a} \approx 1$, the instance is treated as noisy or ambiguous and contributes little to the task loss.

For the task prediction head, NORA mean-pools the contextualized token representations over the target FNE span and passes the resulting span representation to task-specific heads,
\begin{equation*}
    \mathbf{z}_i = \frac{1}{|S_i|}\sum_{t \in S_i} \mathbf{h}_{i,t}, \qquad \mathbf{p}_{i,a} = \mathrm{softmax}\big(f_a(\mathbf{z}_i)\big),
\end{equation*}
where $S_i$ is the set of token positions in the target span, $\mathbf{h}_{i,t}$ the contextual representation of token $t$, and $f_a$ a two-layer MLP for task $a$. We consider tasks $T = \{\texttt{tag}, \texttt{time}, \texttt{scale}, \texttt{sign}\}$, and Appendix~\ref{sec:attr_def} defines the time, scale, and sign labels. Let $\ell_{i,a} = -\log p_{i,a}(y_{i,a})$ denote the cross-entropy loss of instance $i$ for task $a$, set to zero when the label is missing. NORA applies the following noise-modulated loss:
\begin{equation*}
    \tilde{\ell}_{i,a} = (1 - g_{i,a}) \cdot \ell_{i,a} + \hat{\lambda}_{i,a} \cdot g_{i,a}^2.
\end{equation*}

\noindent The first term downweights the raw loss when the instance is estimated to be noisy. The second term regularizes the gate and prevents the trivial solution in which the model sets $g_{i,a}$ close to 1 for every instance. Treating $g_{i,a}$ as a free variable,
\begin{equation*}
\begin{aligned}
    \frac{\partial \tilde{\ell}_{i,a}}{\partial g_{i,a}} &= -\ell_{i,a} + 2\hat{\lambda}_{i,a}\, g_{i,a} = 0, \\
    g^{*}_{i,a} &= \min\big(1, \ell_{i,a} / 2\hat{\lambda}_{i,a}\big),
\end{aligned}
\end{equation*}
since $\tilde{\ell}_{i,a}$ is convex in $g_{i,a}$ and $g_{i,a} \in (0,1)$. Hence, $\hat{\lambda}_{i,a}$ sets how large a loss must be before an instance is ignored. Because loss magnitudes differ widely across tasks and classes and change over the course of training, we set $\hat{\lambda}_{i,a}$ relative to the average loss of the instance's class,
\begin{equation*}
    \hat{\lambda}_{i,a} = \max\big(\lambda_a,\; c \cdot \bar{\ell}_a(y_{i,a})\big),
\end{equation*}
where $\bar{\ell}_a(y)$ is an exponential moving average of the task-$a$ loss over training instances of class $y$, $c$ controls how far above its class average a loss must be before the instance is ignored, and $\lambda_a$ is a lower bound. An instance is fully ignored only when its loss exceeds $2c$ times its class average, so a class in which most instances have high losses is not ignored as a whole. Early in training, losses are high for all instances. With a fixed $\lambda_a$, $g_{i,a}$ would saturate at 1 and most instances would be ignored. Because $\bar{\ell}_a$ follows the loss as it decreases, $\hat{\lambda}_{i,a}$ is large early in training, when the gate tolerates high losses, and shrinks as training proceeds, lowering the loss at which an instance is ignored. Because the optimal gate value depends only on the loss, the gate might simply assign high values to instances with high losses and fail to distinguish hard instances from noisy ones, since both incur high losses. The label embeddings are designed to counter this. If a hard instance is correctly labeled, similar combinations of sentences and labels tend to have low losses and low gate values, so the gate can keep the value of the hard instance low as well and encourage the model to learn from it.

The training objective sums the noise-modulated losses over all tasks $T$ and averages them over a mini-batch of $B$ instances,
\begin{equation*}
    \mathcal{L} = \frac{1}{B}\sum_{i=1}^{B}\sum_{a \in T} \tilde{\ell}_{i,a}.
\end{equation*}

\subsection{Robust Evaluation via Neighborhood Prior-Adjusted KNN Filtering (NPK)}
\label{sec:knn_filtering}

To obtain a more reliable estimate of performance under noisy test labels, we adapt the NPK filtering procedure that identifies high-confidence evaluation subsets.

The NPK procedure is based on the assumption of neighborhood consistency: if an instance is correct, its label should be similar to its neighborhood instances. We implement this idea in three steps.

\noindent \textbf{Semantic feature extraction.} We encode the target sentence alone with a pretrained encoder without fine-tuning and average the final-layer representations of the tokens in the target span. The embeddings are $\ell_2$-normalized so that inner product similarity is equivalent to cosine similarity.

\noindent \textbf{KNN retrieval.} For each sample $x_i$, we retrieve its $k$ nearest neighbors within the same split, including $x_i$ itself, in the semantic embedding space.

\noindent \textbf{Prior-adjusted consistency scoring.} To reduce majority-class bias in the neighborhood counts, we adjust each class count by its global class prior. For class $c$, the prior is
\begin{equation*}
\pi_c = \frac{N_c}{N},
\end{equation*}
where $N_c$ is the number of instances in class $c$ and $N$ is the total number of instances. Let $m_i(c)$ denote the number of neighbors of $x_i$ assigned to class $c$. The prior-adjusted neighborhood score is
\begin{equation*}
q_i(c) = \frac{m_i(c)}{\pi_c + \epsilon},
\end{equation*}
where $\epsilon$ is a small constant that prevents division by zero, and the consistency score for the observed label $y_i$ is
\begin{equation*}
c_i = \frac{q_i(y_i)}{\max_c q_i(c) + \epsilon}.
\end{equation*}

After computing $c_i$ separately for each task for all valid samples, we rank examples in descending order and retain the highest-scoring subsets as cleaner evaluation splits, keeping all instances tied at the cutoff. In our experiments, we report results on the top 90\% subsets. These filtered splits are not assumed to be perfectly noise-free, but they provide a lower-noise evaluation setting that better separates model errors from residual annotation errors in real-world filings.

\noindent \textbf{Relation to SSR.} The prior-adjusted score $c_i$ follows the sample selection of SSR \citep{Feng_2022_BMVC} and its extension NoiseBox \citep{10594806}, which select clean training samples. Because we use it to build evaluation subsets instead, we make three changes. We rank instances and retain a fixed fraction rather than thresholding $c_i$, which controls the subset size. We compute embeddings once with a pretrained encoder without fine-tuning rather than with the model being trained, so that the filter does not depend on the models it evaluates. Finally, we do not relabel instances.

\section{The NORA Dataset}
\label{sec:dataset}
Similar to FiNER-139, FNXL, and EDGAR-CORPUS \citep{loukas-etal-2021-edgar}, \datasetname{} is built from SEC filings. However, our dataset provides 6.3 million sentences with 11.4 million annotated FNEs, compared to 1.1 million sentences in FiNER-139 and 79K sentences in FNXL. Besides tag, time relation, scale, and sign, the dataset retains the fact value, unit of measurement, decimal precision, and axis and member, which we do not use in this study. We also implement a basic noise filter to remove noisy text such as page numbers and TOCs. Appendix~\ref{sec:dataset_appendix} compares \datasetname{} with FiNER-139 and FNXL (Table~\ref{tab:dataset_comparison}) and with datasets in other domains (Table~\ref{tab:label_types}). 

Each instance also includes filing metadata, such as company name, reporting period, and fiscal year, from which time relations can be learned, and the preceding and following sentences. 

The dataset includes 446,338 unique tags and 20 distinct scale values. We apply a frequency filter of 100 to reduce standard tags below the threshold to \texttt{standard\_rare}. All custom tags are reduced to the \texttt{custom} tag. The final tag set contains 3,420 unique tags, more than the 2,794 tags in FNXL.

\paragraph{Deduplication and Data Split} A combined filing, such as a 10-K shared by a parent company and its subsidiaries, is indexed on EDGAR under each co-registrant, so the same document can appear several times. We identify each filing by its SEC accession number, keep one copy of every sentence, and randomly split the filings 80/10/10 into training, validation, and test sets with a fixed seed. All sentences of a filing therefore fall into the same split, and no sentence appears in more than one split. The training, validation, and test sets contain 5,027,025, 626,868, and 627,106 sentences with 9,131,099, 1,139,896, and 1,137,794 FNEs, respectively.

\paragraph{Availability} \datasetname{} is built from public SEC EDGAR filings. We provide a sample of the dataset as supplementary material, and will release the full dataset under the CC BY-NC 4.0 license.

\section{Experiments}
\label{sec:experiments}
\begin{table*}[t]
\centering
\small
\setlength{\tabcolsep}{3pt}
\resizebox{\textwidth}{!}{%
\begin{tabular}{@{}l cc cc cc cc cc@{}}
\toprule
& \multicolumn{2}{c}{\textsc{Tag}} & \multicolumn{2}{c}{\textsc{Time}} & \multicolumn{2}{c}{\textsc{Scale}} & \multicolumn{2}{c}{\textsc{Sign}} & \multicolumn{2}{c}{\textsc{Mean}} \\
\cmidrule(lr){2-3} \cmidrule(lr){4-5} \cmidrule(lr){6-7} \cmidrule(lr){8-9} \cmidrule(lr){10-11}
\textbf{Model} & Acc. & Macro F1 & Acc. & Macro F1 & Acc. & Macro F1 & Acc. & Macro F1 & Acc. & Macro F1 \\
\midrule
FiNER (SEC-BERT)$^{*}$ & \textbf{0.8381} & \textbf{0.7325} & 0.7501 & 0.5819 & \textbf{0.9913} & \underline{0.4661} & \textbf{0.9887} & \textbf{0.8946} & \underline{0.8921} & \underline{0.6688} \\
FiNER (SEC-BERT-SHAPE)$^{*}$ & 0.7997 & 0.6622 & 0.7264 & 0.5598 & 0.9891 & 0.4484 & \underline{0.9878} & 0.8852 & 0.8758 & 0.6389 \\
FNXL (AttentionXML-MASK)$^{*}$ & 0.6136 & 0.2996 & 0.6427 & 0.4443 & 0.9707 & 0.3645 & 0.9816 & 0.8129 & 0.8022 & 0.4803 \\
FNXL (AttentionXML-SHAPE)$^{*}$ & 0.6143 & 0.3029 & 0.6433 & 0.4469 & 0.9746 & 0.3714 & 0.9815 & 0.8158 & 0.8034 & 0.4843 \\
FiNER + SelfMix$^{\dagger}$ & 0.6428 & 0.3017 & 0.7719 & 0.5810 & 0.9788 & 0.3193 & 0.9734 & 0.5282 & 0.8417 & 0.4326 \\
FiNER + SSR$^{\dagger}$ & 0.4929 & 0.3712 & \underline{0.7722} & \underline{0.6086} & 0.9817 & 0.4243 & 0.9518 & 0.7468 & 0.7997 & 0.5377 \\
\midrule
NORA (ours) & \underline{0.8301} & \underline{0.7063} & \textbf{0.8777} & \textbf{0.7468} & \underline{0.9906} & \textbf{0.4667} & 0.9873 & \underline{0.8910} & \textbf{0.9214} & \textbf{0.7027} \\
\bottomrule
\end{tabular}}
\caption{Results on the unfiltered test set. Acc.\ denotes accuracy, which equals micro F1 because each FNE has one label per task. Mean averages each metric over the four tasks. $^{*}$Reads only the target sentence, following the original setting. $^{\dagger}$Applies the method to SEC-BERT with the same input and task heads as NORA, without the gate.}
\label{tab:main_unfiltered}
\end{table*}

\begin{table*}[t]
\centering
\small
\setlength{\tabcolsep}{3pt}
\resizebox{\textwidth}{!}{%
\begin{tabular}{@{}l cc cc cc cc cc@{}}
\toprule
& \multicolumn{2}{c}{\textsc{Tag}} & \multicolumn{2}{c}{\textsc{Time}} & \multicolumn{2}{c}{\textsc{Scale}} & \multicolumn{2}{c}{\textsc{Sign}} & \multicolumn{2}{c}{\textsc{Mean}} \\
\cmidrule(lr){2-3} \cmidrule(lr){4-5} \cmidrule(lr){6-7} \cmidrule(lr){8-9} \cmidrule(lr){10-11}
\textbf{Model} & Acc. & Macro F1 & Acc. & Macro F1 & Acc. & Macro F1 & Acc. & Macro F1 & Acc. & Macro F1 \\
\midrule
FiNER (SEC-BERT)$^{*}$ & \textbf{0.8392} & \textbf{0.7428} & 0.7600 & 0.5941 & \underline{0.9968} & \underline{0.4787} & \underline{0.9916} & \underline{0.9228} & \underline{0.8969} & \underline{0.6846} \\
FiNER (SEC-BERT-SHAPE)$^{*}$ & 0.8002 & 0.6730 & 0.7374 & 0.5722 & 0.9951 & 0.4580 & 0.9908 & 0.9154 & 0.8809 & 0.6547 \\
FNXL (AttentionXML-MASK)$^{*}$ & 0.6149 & 0.3071 & 0.6602 & 0.4592 & 0.9835 & 0.3740 & 0.9849 & 0.8455 & 0.8109 & 0.4965 \\
FNXL (AttentionXML-SHAPE)$^{*}$ & 0.6157 & 0.3103 & 0.6608 & 0.4621 & 0.9874 & 0.3810 & 0.9852 & 0.8507 & 0.8123 & 0.5010 \\
FiNER + SelfMix$^{\dagger}$ & 0.6477 & 0.3119 & 0.7867 & 0.5936 & 0.9936 & 0.3300 & 0.9712 & 0.5281 & 0.8498 & 0.4409 \\
FiNER + SSR$^{\dagger}$ & 0.5424 & 0.4107 & \underline{0.7902} & \underline{0.6357} & 0.9964 & 0.4461 & 0.9765 & 0.8476 & 0.8264 & 0.5850 \\
\midrule
NORA (ours) & \underline{0.8321} & \underline{0.7171} & \textbf{0.8856} & \textbf{0.7589} & \textbf{0.9969} & \textbf{0.4801} & \textbf{0.9922} & \textbf{0.9326} & \textbf{0.9267} & \textbf{0.7222} \\
\bottomrule
\end{tabular}}
\caption{Results on the NPK-filtered test set (top 90\%) when the filter uses RoBERTa embeddings. Acc.\ denotes accuracy, which equals micro F1 because each FNE has one label per task. Mean averages each metric over the four tasks. $^{*}$Reads only the target sentence, following the original setting. $^{\dagger}$Applies the method to SEC-BERT with the same input and task heads as NORA, without the gate.}
\label{tab:main_npk}
\end{table*}
\begin{table*}[t]
\centering
\small
\setlength{\tabcolsep}{3pt}
\resizebox{\textwidth}{!}{%
\begin{tabular}{@{}l ccccc cc cc cc cc cc@{}}
\toprule
& \multicolumn{5}{c}{\textsc{Components}} & \multicolumn{2}{c}{\textsc{Tag}} & \multicolumn{2}{c}{\textsc{Time}} & \multicolumn{2}{c}{\textsc{Scale}} & \multicolumn{2}{c}{\textsc{Sign}} & \multicolumn{2}{c}{\textsc{Mean}} \\
\cmidrule(lr){2-6} \cmidrule(lr){7-8} \cmidrule(lr){9-10} \cmidrule(lr){11-12} \cmidrule(lr){13-14} \cmidrule(lr){15-16}
\textbf{Model} & Meta. & Adj. & Marker & Gate & Label & Acc. & Macro F1 & Acc. & Macro F1 & Acc. & Macro F1 & Acc. & Macro F1 & Acc. & Macro F1 \\
\midrule
FiNER (SEC-BERT) & $\times$ & $\times$ & $\times$ & $\times$ & $\times$ & \textbf{0.8381} & \textbf{0.7325} & 0.7501 & 0.5819 & 0.9913 & \underline{0.4661} & \textbf{0.9887} & \underline{0.8946} & 0.8921 & 0.6688 \\
 & \checkmark & $\times$ & $\times$ & $\times$ & $\times$ & 0.8101 & 0.6855 & 0.8490 & 0.7010 & 0.9904 & 0.4472 & 0.9878 & 0.8866 & 0.9093 \dplus{1.7} & 0.6801 \dplus{1.1} \\
 & \checkmark & \checkmark & $\times$ & $\times$ & $\times$ & 0.8261 & 0.6816 & \underline{0.8777} & 0.7368 & 0.9919 & 0.4384 & 0.9883 & 0.8927 & 0.9210 \dplus{2.9} & 0.6874 \dplus{1.9} \\
 & \checkmark & \checkmark & \checkmark & $\times$ & $\times$ & 0.8251 & 0.6798 & 0.8773 & 0.7370 & \underline{0.9920} & 0.4436 & 0.9884 & 0.8919 & 0.9207 \dplus{2.9} & 0.6881 \dplus{1.9} \\
 & \checkmark & \checkmark & $\times$ & \checkmark & \checkmark & 0.8048 & 0.6544 & 0.8671 & 0.7328 & 0.9899 & 0.4295 & 0.9873 & 0.8875 & 0.9123 \dplus{2.0} & 0.6761 \dplus{0.7} \\
 & \checkmark & \checkmark & \checkmark & \checkmark & $\times$ & \underline{0.8347} & 0.7025 & \textbf{0.8789} & \underline{0.7463} & \textbf{0.9922} & 0.4383 & \textbf{0.9887} & \textbf{0.8947} & \textbf{0.9236} \dplus{3.2} & \underline{0.6955} \dplus{2.7} \\
NORA (ours) & \checkmark & \checkmark & \checkmark & \checkmark & \checkmark & 0.8301 & \underline{0.7063} & \underline{0.8777} & \textbf{0.7468} & 0.9906 & \textbf{0.4667} & 0.9873 & 0.8910 & \underline{0.9214} \dplus{2.9} & \textbf{0.7027} \dplus{3.4} \\
\bottomrule
\end{tabular}}
\caption{Ablation on the unfiltered test set. \checkmark\ marks the components included in each model. Meta.: filing metadata. Adj.: adjacent sentences. Label: label embeddings in the gate. Numbers in parentheses are differences in the mean from FiNER (SEC-BERT) in percentage points.}
\label{tab:ablation}
\end{table*}

\subsection{Experimental Setup}

We implement NORA in PyTorch and initialize the text encoder with \texttt{nlpaueb/sec-bert-base} \citep{loukas2022finer}. Appendix~\ref{sec:training_appendix} lists the hyperparameters and the training procedure. The gate hyperparameters are fixed before training, and Appendix~\ref{sec:c_appendix} reports the sensitivity to $c$. No hyperparameter is tuned on the test set, which is used only once for the final evaluation.

\paragraph{Baselines} We compare NORA with three groups of models. All of them use the same data split, four task heads, batch size, epoch limit, validation schedule, and early stopping as NORA, and their implementation details are given in Appendix~\ref{sec:baseline_appendix}. (1) A vanilla multi-task model uses the same encoder, input, and objective as NORA but without the gate, marker, and label embeddings, which isolates the contribution of the gate from that of joint multi-attribute training. (2) Noisy-label learning methods. SSR \citep{Feng_2022_BMVC} and SelfMix \citep{qiao2022selfmix} separate clean from noisy instances and discard, relabel, or train on model predictions for the noisy ones, whereas NORA keeps every instance and lowers its weight continuously. (3) Financial numeric tagging models. We use SEC-BERT and SEC-BERT-SHAPE from FiNER \citep{loukas2022finer}, which classify the target with a linear layer, and AttentionXML \citep{you2019attentionxml} as used in the second stage of FNXL \citep{sharma2023fnxl}, which labels a given numeral with label-wise attention over a BiLSTM. Following their original settings, these models read only the target sentence. FiNER and FNXL both locate and label numerals in whole sentences on their own datasets, while we specifically focus on tagging numerals. Their reported scores are therefore not comparable to ours, so we retrain these models in our setting.

For NPK filtering, following SSR, we use $k=200$ nearest neighbors and retain the top 90\% of instances. For the main results, the filter uses RoBERTa-base embeddings, which come from a different model family than the SEC-BERT encoder of NORA and most baselines. Appendix~\ref{sec:npk_appendix} reports the results with SEC-BERT embeddings.

\paragraph{Evaluation Metrics}
For all experiments, we report Accuracy and Macro F1 for each task. Because each FNE has one label per task, accuracy equals micro F1, which FiNER and FNXL report alongside macro F1. We emphasize Macro F1 because it weights each class equally under the long-tailed label distributions of FNE tagging. We also report each metric averaged over the four tasks, and the mean Macro F1 is the criterion for early stopping and model selection.

\subsection{Results and Analysis}
We evaluate NORA on the unfiltered test set and on the NPK-filtered subsets (Section~\ref{sec:knn_filtering}).

\paragraph{Unfiltered Test Set}
Table~\ref{tab:main_unfiltered} reports the results on the unfiltered test set. NORA achieves the best overall performance, with the highest mean accuracy (0.9214) and mean Macro F1 (0.7027) over the four tasks, 3.4 points above the strongest baseline, FiNER (SEC-BERT), in mean Macro F1. The largest gain is on Time, where NORA reaches 0.7468 Macro F1 compared with 0.5819 for FiNER (SEC-BERT), which reads only the target sentence. NORA remains competitive on Tag, with a marginal decrease from FiNER (SEC-BERT). We discuss the potential cause in Section~\ref{sec:ablation}, where adding filing metadata substantially improves Time but slightly degrades the other tasks, suggesting a trade-off among tasks. The noisy-label learning baselines and AttentionXML fall far behind, with mean Macro F1 between 0.43 and 0.54. We analyze these results in Appendix~\ref{sec:baseline_analysis}.

\paragraph{Noise-Filtered Test Set}
Table~\ref{tab:main_npk} reports the results on the NPK-filtered test set, where the filter uses RoBERTa embeddings.

The results mostly preserve the main trends from the unfiltered setting. All models score higher than on the unfiltered set, and the ranking by mean Macro F1 is unchanged. \textsc{NORA} again obtains the best mean Accuracy and Macro F1, and the best Accuracy and Macro F1 on Time. These consistent gains suggest that NORA is not simply benefiting from noise in the test labels, but is better capturing context-dependent FNE attributes. The relative performance on Scale remains stable. We note that \textsc{NORA} becomes the best performer on Sign in terms of both Accuracy and Macro F1. With SEC-BERT embeddings, the ranking is also unchanged (Table~\ref{tab:npk_secbert}).
\paragraph{Analysis of the Gate}
To examine whether high gate values correspond to annotation errors, we compute the gates of NORA on the first 200,000 training FNEs and select, for each task, the three instances with the highest and the three with the lowest gate values. A Certified Public Accountant (CPA) reviewed all 24 instances (Appendix~\ref{sec:gate_audit}). Among the 12 high-gate instances, 9 have incorrect gold labels (3 for Tag, 2 for Time, 3 for Scale, and 1 for Sign), whereas all 12 low-gate instances are correctly labeled (one-sided Fisher's exact test, $p = 1.7 \times 10^{-4}$). The Scale errors come from the fact values themselves, such as a percentage not converted to a decimal and \$0.2 million recorded as 200. The remaining high-gate instances, whose labels are correct, reveal room for future improvement. For example, when the text describes a loss, our model tends to predict a negative sign, whereas the sign is positive when the tag itself denotes a loss or an expense.

\begin{table}[t]
\centering
\small
\setlength{\tabcolsep}{4pt}
\begin{tabular}{@{}llcccc@{}}
\toprule
& Label emb. & \textsc{Tag} & \textsc{Time} & \textsc{Scale} & \textsc{Sign} \\
\midrule
\multirow{2}{*}{Spearman $\rho$} & $\times$ & 0.80 & 0.91 & 0.75 & 0.78 \\
 & \checkmark & 0.87 & 0.81 & 0.53 & 0.19 \\
\midrule
\multirow{2}{*}{Kept (\%)} & $\times$ & 99.3 & 95.6 & 99.6 & 100.0 \\
 & \checkmark & 18.4 & 3.2 & 9.7 & 1.1 \\
\bottomrule
\end{tabular}
\caption{Relation between the trained gate value $g$ and the loss-implied optimum $g^{*}$ on the first 200,000 training FNEs. Spearman $\rho$ is computed between $g$ and $g^{*}$. Kept is the percentage of instances with $g^{*} \ge 0.5$ whose gate value is below 0.5.}
\label{tab:gate_optimum}
\end{table}

\paragraph{Gate and the Loss-Implied Optimum}
We compare the trained gate value $g$ with the loss-implied optimum $g^{*}$ on the first 200,000 training FNEs, for the gate with and without label embeddings (Table~\ref{tab:gate_optimum}). Kept denotes the percentage of instances with $g^{*} \ge 0.5$ whose actual gate value is below 0.5, that is, instances that the loss-implied optimum would discard but the gate retains. Without label embeddings, the gate follows the order of $g^{*}$ but keeps its values low (Figure~\ref{fig:gate_distribution} in Appendix~\ref{sec:gate_distribution}), so it retains 95.6--100\% of these instances and fails to discard those whose losses are unusually high for their class. With label embeddings, the gate retains only 1.1--18.4\% of them. At the same time, the rank correlation between $g$ and $g^{*}$ decreases for Time, Scale, and Sign and rises only slightly for Tag. This indicates that the gate no longer depends on the loss alone and may distinguish noisy instances from hard ones, which the CPA review above supports.

\subsection{Ablation Study}
\label{sec:ablation}
We further conduct an ablation study on each component of NORA (Table~\ref{tab:ablation}). Adding filing metadata to FiNER (SEC-BERT) substantially improves Time, raising its Macro F1 from 0.5819 to 0.7010, because the metadata contains the report-period end date and fiscal year from which the time relation is derived. This gain comes with decreases on the other tasks, suggesting a trade-off among tasks. Adding the preceding and following sentences further improves Time and Sign. As expected, the target marker brings no clear improvement on its own (0.6881 vs.\ 0.6874 mean Macro F1), since it mainly serves the gate. It is, however, essential for the gate. Without the marker, the gate with label embeddings lowers the mean Macro F1 to 0.6761, below the model with neither the marker nor the gate (0.6874). With the marker, adding the gate raises the mean Macro F1 from 0.6881 to 0.6955, and adding the label embeddings raises it further to 0.7027. On the NPK-filtered test set, the label embeddings improve Macro F1 on all four tasks over the gate alone, raising the mean from 0.7101 to 0.7222 (Table~\ref{tab:ablation_npk} in Appendix~\ref{sec:npk_appendix}). Only when all components are combined does NORA achieve its best performance. NORA also raises the mean Macro F1 of FinBERT \citep{yang2020finbert} from 0.6477 to 0.6661 (Table~\ref{tab:encoder} in Appendix~\ref{sec:encoder_appendix}).

\section{Conclusion}
\label{sec:conclusion}

We propose \textsc{NORA}, a noise-robust multi-task framework for FNE understanding under noisy labels. \textsc{NORA} uses task-aware instance-specific gates to reduce the influence of unreliable labels during training. We also construct \datasetname{}, a large-scale four-attribute benchmark from SEC filings, and develop an evaluation protocol that filters noisy test labels by neighborhood consistency. On both the unfiltered and noise-filtered test sets, \textsc{NORA} achieves the best overall performance, with the largest gain on \textsc{Time}.


\section{Limitations}
\datasetname{} is built from SEC filings and reflects U.S. reporting practices, so the results may not transfer to other jurisdictions or non-English reports. Our formulation treats the FNE span as given, so errors in span detection could propagate in practice. \datasetname{} also retains attributes beyond the four task targets, such as axis and member, which we leave as auxiliary signals for future work. The predictions of \textsc{NORA} can still be incorrect, especially for rare tags or filings that contain errors, and should support rather than replace expert review. Finally, the neighborhood consistency on which our filtering protocol relies is not equivalent to factual correctness.


\bibliography{custom}


\appendix

\section{Additional Dataset Details for the NORA Dataset}
\label{sec:dataset_appendix}

\paragraph{Comparison with Prior Datasets} Table~\ref{tab:label_types} compares the number of label types in \datasetname{} with entity extraction and extreme multi-label datasets. Table~\ref{tab:dataset_comparison} compares \datasetname{} with FiNER-139 and FNXL.

\begin{table*}[t]
\centering
\small
\begin{tabular}{@{}llr@{}}
\toprule
\textbf{Dataset} & \textbf{Domain} & \textbf{Number of Label Types} \\
\midrule
CoNLL-2003 \citep{tjong-kim-sang-de-meulder-2003-introduction} & Generic & 4 \\
ACE-2005 \citep{doddington-etal-2004-automatic} & Generic & 7 \\
OntoNotes-v5 \citep{pradhan-etal-2012-conll} & Generic & 18 \\
GENIA \citep{10.1093/bioinformatics/btg1023} & Biomedical & 36 \\
EUR-Lex \citep{10.1007/978-3-540-87481-2_4} & Legal & 3,956 \\
\citet{chalkidis2019neural} & Legal & 14 \\
\citet{info10080248} & Financial & 9 \\
FiNER-139 \citep{loukas2022finer} & Financial & 139 \\
FNXL \citep{sharma2023fnxl} & Financial & 2,794 \\
\datasetname{} (ours) & Financial & 3,420 + 9 + 25 + 2 \\
\bottomrule
\end{tabular}
\caption{Number of label types in previous entity extraction and extreme multi-label datasets. For \datasetname{}, the four numbers are the label types of tag, time, scale, and sign.}
\label{tab:label_types}
\end{table*}

\begin{table*}[t]
\centering
\small
\setlength{\tabcolsep}{6pt}
\renewcommand{\arraystretch}{1.2}
\begin{tabular}{@{} l p{0.42\linewidth} p{0.20\linewidth} p{0.20\linewidth} @{}}
\toprule
\textbf{Feature} & \textbf{NORA Dataset} & \textbf{FiNER-139} & \textbf{FNXL} \\
\midrule

\multicolumn{4}{@{}l}{\textit{Dataset Statistics}} \\
\midrule
Size & 6.3M sentences (11.4M FNEs) & 1.1M sentences & 79K sentences \\
Source Filings & 10-K \& 10-Q filings (109,935 filings; 8,226 firms) & 10-K filings & 10-K filings (2.2K firms) \\
Timeframe & 2018--2025 & 2016--2020 & 2019--2021 \\
Tag Inventory & 446K unique tags  (3,420 final tags) & 139 frequent US-GAAP & 2,794 US-GAAP \\
Taxonomy & US-GAAP \& Custom concepts & US-GAAP only & US-GAAP only \\
\midrule

\multicolumn{4}{@{}l}{\textit{Task Formulation \& Context}} \\
\midrule
Target Attributes & Multi-attribute (9 total): \texttt{tag}, \texttt{time},  \texttt{scale}, \texttt{sign}, \texttt{fact}, \texttt{measure}, \texttt{decimals}, \texttt{axis}, \texttt{member} & Tag only & Tag only \\
Provided Context & Target sentence + filing metadata + preceding and following sentences & Target sentence only & Target sentence only \\
\midrule

\multicolumn{4}{@{}l}{\textit{Data Quality}} \\
\midrule
Noise Handling & Semantic filtering; removes irrelevant iXBRL content (e.g., page numbers, TOCs) & Noisy (embedded TOCs and page numbers) & Noisy (inherits standard filing-level noise) \\
Validity & Strict (guarantees $\ge$1 valid tag per instance) & Contains missing or inconsistent tags & May contain missing tags \\

\bottomrule
\end{tabular}
\caption{Comparison of NORA Dataset with FiNER-139 and FNXL. NORA Dataset offers significantly larger scale, comprehensive multi-attribute targets, richer document-level contexts, and cleaner instances.}
\label{tab:dataset_comparison}
\end{table*}

\subsection{Time Relation, Scale, and Sign}
\label{sec:attr_def}
An FNE may correspond to an instant fact, which refers to a specific point in time, or to a duration fact, which spans a time period. For instant facts, the time relation to the filing period is categorized as \textit{past}, \textit{current}, or \textit{future}. For duration facts, the relation can be \textit{past} (spanning within past period), \textit{current} (spanning within current period), \textit{future} (spanning within future period), \textit{past\_current} (from before current period to current), \textit{current\_future} (from current to future), or \textit{past\_future} (from before current period to the future).

The scale attribute specifies the power of ten by which the displayed number should be multiplied to recover its unscaled value. For example, a value reported as 210 in millions has scale 6, corresponding to $210 \times 10^6$. In our data, scale values range from $-12$ to $12$, so we define scale prediction as a 25-class classification task.

The sign attribute is a binary indicator of whether the value should be interpreted as negative with respect to the accounting concept.
 
\subsection{Detailed Schema and Attribute Definitions}

Each instance in \datasetname{} is organized into four components. The \textit{document} field stores filing-level metadata such as the reporting entity name, CIK, and reporting period; \textit{context} stores the target sentence containing the numeric mention; \textit{targets} identifies the character-aligned text span to be tagged; and \textit{golds} stores the aligned XBRL attribute values for that target. Table~\ref{tab:attribute_description_appendix} summarizes the attribute definitions used in \datasetname{}.
\begin{table*}[t]
\centering
\small
\begin{tabular}{lp{12cm}}
\hline
\textbf{Attribute} & \textbf{Description} \\
\hline
\multicolumn{2}{l}{\textit{Attributes used in the task definition}} \\
\hline
\texttt{tag} & Custom tags are generically labeled as \texttt{custom}. Standard tags that occur fewer than 100 times within the original training split are classified as \texttt{standard\_rare}. Conversely, standard tags that appear at least 100 times retain their exact original names. Standard namespaces include \texttt{us-gaap}, as well as others like \texttt{dei} and \texttt{srt}. \\
\texttt{time} & This attribute is initially categorized into two period types: \textit{instant} and \textit{period}. It is then further classified into simplified timeframes, such as \textit{past}, \textit{current}, \textit{future}, or a combination of these, e.g., \textit{past\_current} or \textit{current\_future}. The specific accounting period depends on the document type, with 10-K filings corresponding to a one-year period and 10-Q filings corresponding to a three-month period. \\
\texttt{scale} & This represents the scaling factors for the numerical value. \\
\texttt{sign} & A binary indicator of whether the numerical value should be interpreted as negative with respect to the accounting concept. \\
\hline
\multicolumn{2}{l}{\textit{Additional attributes not used as task targets in this paper}} \\
\hline
\texttt{fact} & This represents the factual value, which is formatted as a float.   \\
\texttt{measure} & The unit of measurement. \\
\texttt{decimals} & The declared decimal precision. \\
\texttt{axis} (list) & A list of dimensions classifying the fact. Each item contains \texttt{amseq}, which links the dimension to its member, and \texttt{axis}, which stores the axis name. \\
\texttt{member} (list) & A list of members associated with the fact. Each item contains \texttt{amseq}, which links the member to its axis, and \texttt{member}, which stores the member name. \\
\hline
\end{tabular}
\caption{XBRL target attributes in \datasetname{}, divided into attributes used in the paper's task definition and additional attributes retained in the dataset.}
\label{tab:attribute_description_appendix}
\end{table*}

Figure~\ref{fig:task1_dataset_example} illustrates the JSONL instances in \datasetname{}. The key fields in each instance are summarized below. 

\begin{figure*}[t]
\centering
\lstset{basicstyle=\ttfamily\small,breaklines=true,breakatwhitespace=false,columns=fullflexible,keepspaces=true,showstringspaces=false}\begin{lstlisting}
{
  "job_id": "000032019324000123-00030",
  "document": {
    "company_name": "Apple Inc.",
    "cik": "0000320193",
    "document_type": "10-K",
    "period_end_date": "2024-09-28",
    "fiscal_year": "2024",
    "period_focus": "FY",
    "current_fiscal_year_end": "--09-28",
    "document_link": "https://www.sec.gov/ix?doc=/Archives/edgar/data/0000320193/000032019324000123/aapl-20240928.htm"
  },
  "context": {
    "context_p": "The Company uses net proceeds from the commercial paper program for general corporate purposes, including dividends and share repurchases.",
    "context_t": "As of September 28, 2024 and September 30, 2023, the Company had $10.0 billion and $6.0 billion of commercial paper outstanding, respectively, with maturities generally less than nine months.",
    "context_n": "The weighted-average interest rate of the Company's commercial paper was 5.00% and 5.28% as of September 28, 2024 and September 30, 2023, respectively."
  },
  "targets": [
    {
      "seq_id": 0,
      "start_pos": 66,
      "end_pos": 70,
      "text": "10.0",
      "attribute": ["tag", "fact", "time", "axis", "member", "measure", "decimals", "scale"]
    },
    {
      "seq_id": 1,
      "start_pos": 84,
      "end_pos": 87,
      "text": "6.0",
      "attribute": ["tag", "fact", "time", "axis", "member", "measure", "decimals", "scale"]
    }
  ],
  "golds": [
    {
      "seq_id": 0,
      "value": [
        "us-gaap:CommercialPaper",
        10000000000.0,
        "instant: 2024-09-28",
        [{"amseq": 0, "axis": "us-gaap:ShortTermDebtTypeAxis"}],
        [{"amseq": 0, "member": "us-gaap:CommercialPaperMember"}],
        "iso4217:USD",
        "-8",
        "9"
      ]
    },
    {
      "seq_id": 1,
      "value": [
        "us-gaap:CommercialPaper",
        6000000000.0,
        "instant: 2023-09-30",
        [{"amseq": 0, "axis": "us-gaap:ShortTermDebtTypeAxis"}],
        [{"amseq": 0, "member": "us-gaap:CommercialPaperMember"}],
        "iso4217:USD",
        "-8",
        "9"
      ]
    }
  ]
}
\end{lstlisting}
\caption{Example JSONL data instance from \datasetname{}.}
\label{fig:task1_dataset_example}
\end{figure*}

\begin{description}
    \item[\texttt{document}:] Contains essential metadata about the financial filing, such as the company name, document type (e.g., 10-K), fiscal year, and a direct URL to the source iXBRL file.
    \item[\texttt{context}:] Provides the local textual surroundings for the AI model to read. It includes:
    \begin{itemize}
        \item \texttt{context\_t}: The \textbf{target} sentence that contains the numbers being tagged.
        \item \texttt{context\_p}: The \textbf{preceding} sentence.
        \item \texttt{context\_n}: The \textbf{following} sentence.
    \end{itemize}
    \item[\texttt{targets}:] Identifies the exact word or number sequence in the target sentence that requires tagging. It uses character-level start and end positions to pinpoint the text (e.g., locating "10.0" and "6.0") and lists the attributes the model is expected to predict.
    \item[\texttt{golds}:] The "ground truth" or correct answers for training and testing. It provides the exact taxonomy tags, the unscaled numerical facts, the time periods, and the measurement units, matched perfectly to the target IDs.
\end{description}

\paragraph{Additional Statistics for \datasetname{}}
Table~\ref{tab:tag_distribution} lists the ten most frequent tags and their proportions across the train, validation, and test splits. Table~\ref{tab:time_category_distribution} summarizes the distribution of time-relation categories across the same splits.

\begin{table*}[t]
\centering
\normalsize
\setlength{\tabcolsep}{6pt}
\renewcommand{\arraystretch}{1.22}
\begin{tabular}{@{}>{\raggedright\arraybackslash}p{0.68\textwidth}rrr@{}}
\toprule
\textbf{Tag Name} & \textbf{Train} & \textbf{Validation} & \textbf{Test} \\
\midrule
us-gaap:DebtInstrumentInterest\newline RateStatedPercentage & 1.84\% & 1.85\% & 1.72\% \\
\addlinespace[2pt]
us-gaap:DebtInstrumentBasisSpread\newline OnVariableRate1 & 1.50\% & 1.55\% & 1.48\% \\
\addlinespace[2pt]
us-gaap:DebtInstrument\newline FaceAmount & 1.43\% & 1.45\% & 1.40\% \\
\addlinespace[2pt]
us-gaap:LineOfCreditFacility\newline MaximumBorrowingCapacity & 1.37\% & 1.40\% & 1.34\% \\
\addlinespace[2pt]
us-gaap:AllocatedShareBased\newline CompensationExpense & 1.17\% & 1.14\% & 1.13\% \\
\addlinespace[2pt]
us-gaap:ConcentrationRisk\newline Percentage1 & 0.97\% & 0.98\% & 0.98\% \\
\addlinespace[2pt]
dei:EntityCommonStock\newline SharesOutstanding & 0.88\% & 0.88\% & 0.88\% \\
\addlinespace[2pt]
us-gaap:AmortizationOf\newline IntangibleAssets & 0.74\% & 0.74\% & 0.74\% \\
\addlinespace[2pt]
us-gaap:AntidilutiveSecuritiesExcluded\newline FromComputationOfEarningsPerShareAmount & 0.72\% & 0.70\% & 0.73\% \\
\addlinespace[2pt]
us-gaap:EffectiveIncomeTaxRate\newline ContinuingOperations & 0.66\% & 0.68\% & 0.67\% \\
\bottomrule
\end{tabular}
\caption{Distribution of the ten most frequent tags across the splits of \datasetname{}. Percentages are computed over all FNEs, including those labeled \texttt{custom} (26.8\% of the training set) and \texttt{standard\_rare} (1.2\%).}
\label{tab:tag_distribution}
\end{table*}

\begin{table*}[t]
\centering
\begin{tabular}{lrrrrr}
\toprule
\multicolumn{1}{c}{\multirow{2}{*}{\textbf{Time}}} & \multicolumn{1}{c}{\multirow{2}{*}{\textbf{Train}}} & \multicolumn{1}{c}{\multirow{2}{*}{\textbf{Validation}}} & \multicolumn{1}{c}{\multirow{2}{*}{\textbf{Test}}} & \multicolumn{2}{c}{\textbf{All}} \\
\cmidrule(lr){5-6}
& & & & \multicolumn{1}{c}{\textbf{\# of}} & \multicolumn{1}{c}{\textbf{\%}} \\
\midrule
instant; past & 18.80\% & 18.80\% & 18.96\% & 2,141,143 & 18.82\% \\
instant; current & 26.68\% & 26.48\% & 26.54\% & 3,032,142 & 26.65\% \\
instant; future & 2.07\% & 2.07\% & 2.05\% & 235,407 & 2.07\% \\
\midrule
\multicolumn{1}{c}{Total Instant} & 47.55\% & 47.35\% & 47.55\% & 5,408,692 & 47.53\% \\
\midrule
duration; past & 24.12\% & 24.33\% & 24.33\% & 2,749,634 & 24.16\% \\
duration; current & 17.15\% & 16.98\% & 16.91\% & 1,946,571 & 17.11\% \\
duration; future & 1.42\% & 1.43\% & 1.43\% & 162,239 & 1.43\% \\
duration; past\_current & 9.65\% & 9.80\% & 9.67\% & 1,100,328 & 9.67\% \\
duration; past\_future & 0.06\% & 0.07\% & 0.06\% & 7,252 & 0.06\% \\
duration; current\_future & 0.04\% & 0.04\% & 0.04\% & 4,895 & 0.04\% \\
\midrule
\multicolumn{1}{c}{Total Period} & 52.45\% & 52.65\% & 52.45\% & 5,970,919 & 52.47\% \\
\midrule
\multicolumn{1}{c}{Total Time Count} & 9,108,297 & 1,136,972 & 1,134,342 & 11,379,611 & 100.00\% \\
\bottomrule
\end{tabular}
\caption{Distribution of Time Category in \datasetname{}}
\label{tab:time_category_distribution}
\end{table*}

\section{Training Details}
\label{sec:training_appendix}
For the gate, we set $\lambda_a = 1$ for all tasks, $c = 5$, and the moving-average momentum to 0.99. These values are fixed before training, and Figure~\ref{fig:c_ablation} reports the sensitivity to $c$.

We optimize all models with AdamW and a weight decay of $0.01$. We use a learning rate of $1 \times 10^{-5}$ for the pretrained encoder, within the ranges used by \citet{liu2019robertarobustlyoptimizedbert} and \citet{loukas2022finer}, and $1 \times 10^{-4}$ for the randomly initialized components (task heads, gate, marker, and label embeddings) \citep{howard-ruder-2018-universal}. The learning rate increases linearly over the first 6\% of steps and then decays linearly to zero \citep{liu2019robertarobustlyoptimizedbert}. We train with a batch size of 64 for at most 5 epochs and evaluate on the full validation set every 10,000 steps. Early stopping is applied from the third epoch, with a patience of three evaluations on the mean macro F1 over the four tasks, and we keep the checkpoint with the best validation score. The SEC-BERT and FinBERT encoders have about 110M parameters each. We train each model on a single NVIDIA GeForce RTX 4090 or RTX 5090 GPU, and one full training run takes about two days, so the about 15 runs in this paper take about 720 GPU hours in total. Each reported result comes from a single run with a fixed random seed.

\section{Baseline Implementation Details}
\label{sec:baseline_appendix}
All baselines share NORA's data split, batch size of 64, limit of 5 epochs, evaluation on the full validation set every 10,000 steps, and early stopping rule. We follow the original papers and code, and list below where our implementation differs and why.

\paragraph{FiNER} SEC-BERT and SEC-BERT-SHAPE read only the target sentence and classify the target with a linear layer, as in \citet{loukas2022finer}. We attach one linear head per task. For SEC-BERT-SHAPE, each numeral is replaced with its shape pseudo-token, and a numeral whose shape is not in the vocabulary is kept unchanged.

\paragraph{AttentionXML} We follow the original code and the EUR-Lex configuration \citep{Bhatia16} used by FNXL, with frozen 300-dimensional GloVe embeddings \citep{pennington-etal-2014-glove}, a BiLSTM with 256 hidden units, dropout of 0.5, label-wise attention, a 256-unit fully connected layer, binary cross-entropy, and Adam with a learning rate of $10^{-3}$ and the original gradient clipping. As in FNXL, the model reads only the target sentence, and other numerals are replaced with [MASK] or shape tokens. The four tasks share the embeddings and the BiLSTM, and each task has its own attention and output layers. The binary cross-entropy is summed over classes so that the tag task is not outweighed by tasks with fewer classes. Stochastic weight averaging \citep{Izmailov2018AveragingWL} starts from the second epoch instead of the tenth of 30 epochs. FNXL validates on 200 training instances with nDCG@5, which we replace with our validation set and macro F1.

\paragraph{SSR} We use $K = 200$ from the original code, $\theta_r = 0.95$ for real-world datasets as in the paper, and $\theta_s = 1$, and keep the class-balanced vote of \citet{Feng_2022_BMVC}. Relabeling and selection run at step 0 and at the start of every epoch, separately for each task, so an instance can be clean for one task and noisy for another.

\paragraph{SelfMix} We use the instance-dependent-noise setting of \citet{qiao2022selfmix} ($\lambda_p = 0$) with $\tau = 0.5$, $T = 0.5$, $\alpha = 0.75$, and $\lambda_r = 0.3$. The cross-entropy warm-up lasts 42,801 steps, the same as the learning-rate warm-up, and samples are reselected at every epoch. Selection, sharpening, and EmbMix are applied separately for each task with a per-instance mask, and EmbMix mixes the target span representation instead of [CLS], because our task heads read the span. Where the released code differs from the paper, we follow the paper. Selection uses the cross-entropy loss rather than a loss computed after two softmax operations, class regularization divides by the standard deviation rather than the variance, and the pseudo loss uses the predicted class.

\section{Analysis of Baselines}
\label{sec:baseline_analysis}
The noisy-label learning baselines were evaluated on label spaces of at most 100 classes, and both struggle with the 3,420 tags of \datasetname{}. SSR divides each neighbor vote by the class prior, so an instance of a frequent tag rarely wins the vote for its own label when rare tags appear among its neighbors. SSR keeps only 13.9\% of the training instances for Tag at the start of training and 41.8\% after the first epoch, compared with 44.4\% and 81.0\% for Time, and its Tag accuracy (0.4929) is the lowest among all models. SelfMix does not discard the instances it judges as noisy but trains them on the sharpened predictions of the model, so labels of rare classes may be replaced by those of frequent classes. Its Sign predictions collapse to the majority class, with an accuracy of 0.9734 but a Macro F1 of only 0.5282. AttentionXML labels each target separately, as the BERT-based models do, but it can identify the target only as the numeral left unmasked in the sentence, and its frozen GloVe embeddings and tokenization discard symbols such as \$, \%, and parentheses.

\section{NPK Implementation Details}
\label{sec:npk_appendix}
We set $\epsilon = 10^{-10}$. Because $c_i$ is computed from integer vote counts and a fixed prior, it takes only a few distinct values. All instances tied at the 90\% cutoff are retained, so the retained fraction can exceed 90\% (91.05\% for scale with SEC-BERT). Table~\ref{tab:npk_secbert} reports the results when the filter uses SEC-BERT embeddings.

\begin{table*}[t]
\centering
\small
\setlength{\tabcolsep}{3pt}
\resizebox{\textwidth}{!}{%
\begin{tabular}{@{}l cc cc cc cc cc@{}}
\toprule
& \multicolumn{2}{c}{\textsc{Tag}} & \multicolumn{2}{c}{\textsc{Time}} & \multicolumn{2}{c}{\textsc{Scale}} & \multicolumn{2}{c}{\textsc{Sign}} & \multicolumn{2}{c}{\textsc{Mean}} \\
\cmidrule(lr){2-3} \cmidrule(lr){4-5} \cmidrule(lr){6-7} \cmidrule(lr){8-9} \cmidrule(lr){10-11}
\textbf{Model} & Acc. & Macro F1 & Acc. & Macro F1 & Acc. & Macro F1 & Acc. & Macro F1 & Acc. & Macro F1 \\
\midrule
FiNER (SEC-BERT)$^{*}$ & \textbf{0.8398} & \textbf{0.7430} & 0.7633 & 0.5973 & 0.9975 & \underline{0.4836} & \underline{0.9925} & \underline{0.9303} & \underline{0.8983} & \underline{0.6886} \\
FiNER (SEC-BERT-SHAPE)$^{*}$ & 0.8011 & 0.6733 & 0.7418 & 0.5760 & 0.9959 & 0.4624 & 0.9919 & 0.9239 & 0.8827 & 0.6589 \\
FNXL (AttentionXML-MASK)$^{*}$ & 0.6171 & 0.3075 & 0.6670 & 0.4639 & 0.9848 & 0.3748 & 0.9861 & 0.8549 & 0.8138 & 0.5003 \\
FNXL (AttentionXML-SHAPE)$^{*}$ & 0.6178 & 0.3107 & 0.6677 & 0.4668 & 0.9887 & 0.3827 & 0.9865 & 0.8609 & 0.8152 & 0.5053 \\
FiNER + SelfMix$^{\dagger}$ & 0.6503 & 0.3128 & 0.7933 & 0.5989 & 0.9955 & 0.3309 & 0.9711 & 0.5280 & 0.8526 & 0.4427 \\
FiNER + SSR$^{\dagger}$ & 0.5436 & 0.4116 & \underline{0.7986} & \underline{0.6471} & \textbf{0.9981} & 0.4509 & 0.9858 & 0.8964 & 0.8315 & 0.6015 \\
\midrule
NORA (ours) & \underline{0.8329} & \underline{0.7175} & \textbf{0.8888} & \textbf{0.7624} & \underline{0.9979} & \textbf{0.4852} & \textbf{0.9938} & \textbf{0.9446} & \textbf{0.9284} & \textbf{0.7274} \\
\bottomrule
\end{tabular}}
\caption{Results on the NPK-filtered test set (top 90\%) when the filter uses SEC-BERT embeddings. Acc.\ denotes accuracy, which equals micro F1 because each FNE has one label per task. Mean averages each metric over the four tasks. $^{*}$Reads only the target sentence, following the original setting. $^{\dagger}$Applies the method to SEC-BERT with the same input and task heads as NORA, without the gate.}
\label{tab:npk_secbert}
\end{table*}

Table~\ref{tab:ablation_npk} reports the ablation on the NPK-filtered test set with the RoBERTa filter.

\begin{table*}[t]
\centering
\small
\setlength{\tabcolsep}{3pt}
\resizebox{\textwidth}{!}{%
\begin{tabular}{@{}l ccccc cc cc cc cc cc@{}}
\toprule
& \multicolumn{5}{c}{\textsc{Components}} & \multicolumn{2}{c}{\textsc{Tag}} & \multicolumn{2}{c}{\textsc{Time}} & \multicolumn{2}{c}{\textsc{Scale}} & \multicolumn{2}{c}{\textsc{Sign}} & \multicolumn{2}{c}{\textsc{Mean}} \\
\cmidrule(lr){2-6} \cmidrule(lr){7-8} \cmidrule(lr){9-10} \cmidrule(lr){11-12} \cmidrule(lr){13-14} \cmidrule(lr){15-16}
\textbf{Model} & Meta. & Adj. & Marker & Gate & Label & Acc. & Macro F1 & Acc. & Macro F1 & Acc. & Macro F1 & Acc. & Macro F1 & Acc. & Macro F1 \\
\midrule
FiNER (SEC-BERT) & $\times$ & $\times$ & $\times$ & $\times$ & $\times$ & \textbf{0.8392} & \textbf{0.7428} & 0.7600 & 0.5941 & 0.9968 & \underline{0.4787} & 0.9916 & 0.9228 & 0.8969 & 0.6846 \\
 & \checkmark & $\times$ & $\times$ & $\times$ & $\times$ & 0.8116 & 0.6969 & 0.8576 & 0.7140 & 0.9967 & 0.4601 & 0.9910 & 0.9173 & 0.9142 \dplus{1.7} & 0.6971 \dplus{1.2} \\
 & \checkmark & \checkmark & $\times$ & $\times$ & $\times$ & 0.8255 & 0.6907 & 0.8849 & 0.7474 & \underline{0.9970} & 0.4490 & \underline{0.9917} & 0.9247 & 0.9248 \dplus{2.8} & 0.7030 \dplus{1.8} \\
 & \checkmark & \checkmark & \checkmark & $\times$ & $\times$ & 0.8239 & 0.6884 & 0.8844 & 0.7474 & \textbf{0.9971} & 0.4547 & 0.9914 & 0.9217 & 0.9242 \dplus{2.7} & 0.7031 \dplus{1.8} \\
 & \checkmark & \checkmark & $\times$ & \checkmark & \checkmark & 0.8068 & 0.6660 & 0.8758 & 0.7473 & 0.9963 & 0.4420 & 0.9916 & \underline{0.9253} & 0.9176 \dplus{2.1} & 0.6952 \dplus{1.1} \\
 & \checkmark & \checkmark & \checkmark & \checkmark & $\times$ & \underline{0.8338} & 0.7110 & \textbf{0.8860} & \underline{0.7567} & \underline{0.9970} & 0.4493 & 0.9916 & 0.9235 & \textbf{0.9271} \dplus{3.0} & \underline{0.7101} \dplus{2.6} \\
NORA (ours) & \checkmark & \checkmark & \checkmark & \checkmark & \checkmark & 0.8321 & \underline{0.7171} & \underline{0.8856} & \textbf{0.7589} & 0.9969 & \textbf{0.4801} & \textbf{0.9922} & \textbf{0.9326} & \underline{0.9267} \dplus{3.0} & \textbf{0.7222} \dplus{3.8} \\
\bottomrule
\end{tabular}}
\caption{Ablation on the NPK-filtered test set (RoBERTa filter). \checkmark\ marks the components included in each model. Meta.: filing metadata. Adj.: adjacent sentences. Label: label embeddings in the gate. Numbers in parentheses are differences in the mean from FiNER (SEC-BERT) in percentage points.}
\label{tab:ablation_npk}
\end{table*}

\section{Sensitivity to \texorpdfstring{$c$}{c}}
\label{sec:c_appendix}
\begin{figure}[t]
\centering
\includegraphics[width=\columnwidth]{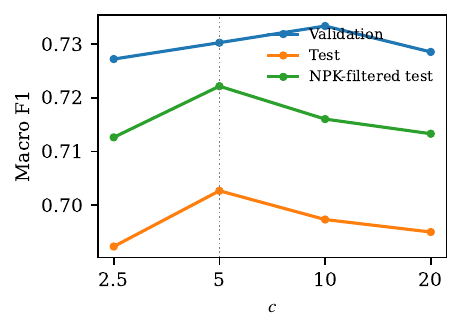}
\caption{Mean macro F1 over the four tasks of NORA with different values of $c$ on the validation, test, and NPK-filtered test (RoBERTa filter) sets. The dotted line marks $c = 5$, the value used in the main results.}
\label{fig:c_ablation}
\end{figure}

\begin{figure*}[t]
\centering
\begin{minipage}{0.48\textwidth}\centering\includegraphics[width=\linewidth]{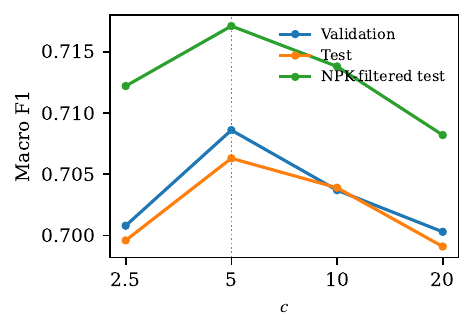}\\(a) Tag\end{minipage}\hfill
\begin{minipage}{0.48\textwidth}\centering\includegraphics[width=\linewidth]{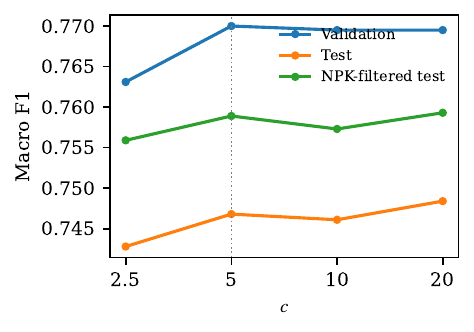}\\(b) Time\end{minipage}\\[6pt]
\begin{minipage}{0.48\textwidth}\centering\includegraphics[width=\linewidth]{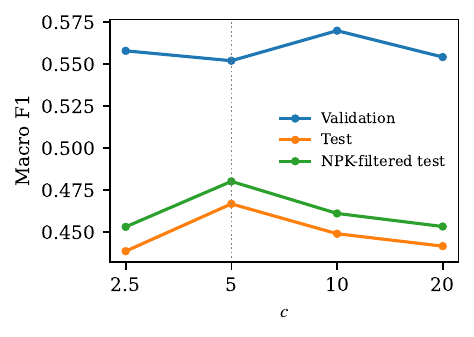}\\(c) Scale\end{minipage}\hfill
\begin{minipage}{0.48\textwidth}\centering\includegraphics[width=\linewidth]{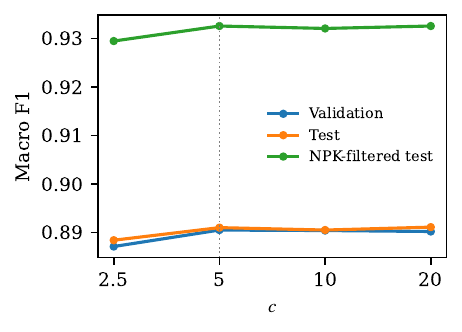}\\(d) Sign\end{minipage}
\caption{Macro F1 of each task of NORA with different values of $c$ on the validation, test, and NPK-filtered test (RoBERTa filter) sets. The dotted line marks $c = 5$.}
\label{fig:c_tasks}
\end{figure*}

Figure~\ref{fig:c_ablation} shows the mean macro F1 over the four tasks for different values of $c$, and Figure~\ref{fig:c_tasks} shows the macro F1 of each task.

\section{Results with Another Encoder}
\label{sec:encoder_appendix}
Table~\ref{tab:encoder} reports the results with SEC-BERT and FinBERT \citep{yang2020finbert} as the encoder.

\begin{table*}[t]
\centering
\small
\setlength{\tabcolsep}{3pt}
\resizebox{\textwidth}{!}{%
\begin{tabular}{@{}l cc cc cc cc cc@{}}
\toprule
& \multicolumn{2}{c}{\textsc{Tag}} & \multicolumn{2}{c}{\textsc{Time}} & \multicolumn{2}{c}{\textsc{Scale}} & \multicolumn{2}{c}{\textsc{Sign}} & \multicolumn{2}{c}{\textsc{Mean}} \\
\cmidrule(lr){2-3} \cmidrule(lr){4-5} \cmidrule(lr){6-7} \cmidrule(lr){8-9} \cmidrule(lr){10-11}
\textbf{Model} & Acc. & Macro F1 & Acc. & Macro F1 & Acc. & Macro F1 & Acc. & Macro F1 & Acc. & Macro F1 \\
\midrule
SEC-BERT (vanilla) & 0.8261 & 0.6816 & \textbf{0.8777} & 0.7368 & \textbf{0.9919} & 0.4384 & \textbf{0.9883} & \textbf{0.8927} & 0.9210 & 0.6874 \\
SEC-BERT + NORA & \textbf{0.8301} & \textbf{0.7063} & \textbf{0.8777} & \textbf{0.7468} & 0.9906 & \textbf{0.4667} & 0.9873 & 0.8910 & \textbf{0.9214} & \textbf{0.7027} \\
\midrule
FinBERT (vanilla) \citep{yang2020finbert} & 0.7779 & 0.5898 & 0.8577 & 0.7058 & \textbf{0.9906} & 0.4152 & \textbf{0.9873} & 0.8799 & 0.9034 & 0.6477 \\
FinBERT + NORA & \textbf{0.7948} & \textbf{0.6353} & \textbf{0.8628} & \textbf{0.7163} & 0.9879 & \textbf{0.4296} & 0.9864 & \textbf{0.8831} & \textbf{0.9080} & \textbf{0.6661} \\
\bottomrule
\end{tabular}}
\caption{Results with different encoders on the unfiltered test set. Vanilla uses the same input and task heads as NORA without the gate, marker, and label embeddings. Bold marks the better result within each encoder.}
\label{tab:encoder}
\end{table*}

\section{Gate Audit Details}
\label{sec:gate_audit}
Table~\ref{tab:gate_audit} lists the 24 training instances reviewed by the CPA, the three with the highest and the three with the lowest gate values for each task. Tables~\ref{tab:audit-tag-high-1}--\ref{tab:audit-sign-low-3} give the context of each instance.

\begin{table*}[t]
\centering
\scriptsize
\setlength{\tabcolsep}{3pt}
\renewcommand{\arraystretch}{1.15}
\begin{tabular}{@{}l l r p{1.5cm} p{4.2cm} p{3.0cm} p{4.4cm}@{}}
\toprule
\textbf{Task} & \textbf{Gate} & \textbf{$g$} & \textbf{Target} & \textbf{Gold label} & \textbf{Corrected label} & \textbf{Reason} \\
\midrule
Tag & High & 1.0000 & \$6.41 & \url{us-gaap:ShareBasedCompensationArrangementByShareBasedPaymentAwardEquityInstrumentsOtherThanOptionsGrantsInPeriodWeightedAverageGrantDateFairValue} & \url{custom} or \url{us-gaap:BusinessAcquisitionSharePrice} & Minimum price for acquiring the Floating Shares, not the grant-date fair value of RSUs. \\
Tag & High & 1.0000 & \$4.0 million & \url{custom} & \url{us-gaap:AllocatedShareBasedCompensationExpense} & PSU expense is already identified by the award-type axis, so the standard element suffices. \\
Tag & High & 1.0000 & 1\% & \url{us-gaap:EffectiveIncomeTaxRateContinuingOperations} & \url{custom} & Excise tax rate on share repurchases under the Inflation Reduction Act, not the effective income tax rate. \\
Tag & Low & 0.0010 & 11 & \url{custom} & -- & Maximum leverage ratio in a debt covenant. \\
Tag & Low & 0.0010 & eleven & \url{custom} & -- & Number of lenders. \\
Tag & Low & 0.0010 & four & \url{custom} & -- & Number of vesting tranches. \\
\midrule
Time & High & 1.0000 & 32,000,000 & instant; past & instant; current & A plan adopted in the past but currently effective, so the shares available for grant are reported for the current period. \\
Time & High & 1.0000 & \$58.5 million & period; past & -- & Contract liabilities recognized as revenue in a past period. \\
Time & High & 1.0000 & \$7.5 million & instant; current & period; past & Earnout cash received in the first quarter of 2021, an amount for a past period. \\
Time & Low & 0.0004 & \$465,860 & instant; current & -- & The date in the sentence is the end of the current period. \\
Time & Low & 0.0004 & \$453,629 & instant; current & -- & The date in the sentence is the end of the current period. \\
Time & Low & 0.0004 & \$1.71 billion & instant; current & -- & The date in the sentence is the end of the current period. \\
\midrule
Scale & High & 1.0000 & 2.454\% & 0 & $-2$ & The fact is recorded as 2.454 instead of 0.02454. \\
Scale & High & 1.0000 & 1.123\% & 0 & $-2$ & The fact is recorded as 1.123 instead of 0.01123. \\
Scale & High & 1.0000 & \$0.2 million & 3 & 6 & The fact is recorded as 200 instead of 200,000. \\
Scale & Low & 0.0000 & \$2.4 trillion & 12 & -- & Trillion corresponds to a scale of 12. \\
Scale & Low & 0.0000 & \$1.5 trillion & 12 & -- & Trillion corresponds to a scale of 12. \\
Scale & Low & 0.0000 & \$1.3 trillion & 12 & -- & Trillion corresponds to a scale of 12. \\
\midrule
Sign & High & 1.0000 & \$72.1 million & positive & negative, with \url{us-gaap:VariableInterestEntityInitialConsolidationGainOrLoss} & INN is a variable interest entity, so the initial-consolidation gain-or-loss element fits better than the acquisition-cost element. It has a credit balance, so the loss is negative. \\
Sign & High & 1.0000 & \$2.2 million & positive & -- & \url{us-gaap:LossOnDerivativeInstrumentsPretax} is a loss element, so a positive value is correct. \\
Sign & High & 1.0000 & \$68,000 & positive & -- & \url{us-gaap:LossOnDerivativeInstrumentsPretax} is a loss element, so a positive value is correct. \\
Sign & Low & 0.0003 & 6.40\% & positive & -- & Interest rate. \\
Sign & Low & 0.0003 & 2.40\% & positive & -- & Interest rate. \\
Sign & Low & 0.0003 & 0.125\% & positive & -- & Interest rate. \\
\bottomrule
\end{tabular}
\caption{CPA review of the 24 training instances with the highest and lowest gate values for each task. A dash in the Corrected label column means that the CPA judged the gold label correct.}
\label{tab:gate_audit}
\end{table*}

\section{Gate Value Distribution}
\label{sec:gate_distribution}
\begin{figure*}[t]
\centering
\includegraphics[width=0.8\textwidth]{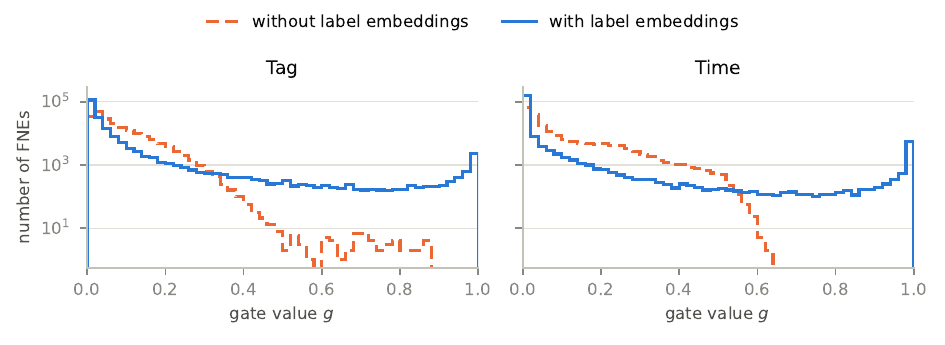}
\caption{Distribution of gate values for Tag and Time on the first 200,000 training FNEs, with and without label embeddings in the gate. The vertical axis is on a log scale.}
\label{fig:gate_distribution}
\end{figure*}

Figure~\ref{fig:gate_distribution} shows the distribution of gate values on the first 200,000 training FNEs for the gate with and without label embeddings.

\begin{table*}[p]
\centering
\small
\setlength{\tabcolsep}{2pt}
\begin{tabular}{@{} >{\bfseries}p{0.14\textwidth} >{\raggedright\arraybackslash}p{0.82\textwidth} @{}}
\toprule
Field & Value \\
\midrule
Context & \textbf{Target Sentence:} Pursuant to the Amended Plan of Arrangement, upon the occurrence or waiver (at the discretion of Canopy Growth) of the Triggering Event (the “Triggering Event Date”), Canopy Growth will, subject to the satisfaction or waiver of certain closing conditions set out in the Amending Agreement: (i) acquire all of the issued and outstanding Fixed Shares (following the mandatory conversion of the Fixed Multiple Shares into Fixed Shares) on the basis of 0.3048 of a common share of Canopy Growth (each whole common share, a “Canopy Growth Share”) for each Fixed Share held (the “Fixed Exchange Ratio”) at the time of the acquisition of the Fixed Shares (the “Acquisition Time”), subject to adjustment in accordance with the terms of the Amended Plan of Arrangement (the “Canopy Call Option”) and (ii) have the right (but not the obligation) (the “Floating Call Option”), exercisable for a period of 30 days following the Triggering Event Date to acquire all of the issued and outstanding Floating Shares at a price to be determined based upon the 30 day volume-weighted average trading price of the Floating Shares, subject to a minimum price of \$[6.41], as may be adjusted in accordance with the terms of the Amended Plan of Arrangement, to be payable, at the option of Canopy Growth, in cash, Canopy Growth Shares, or a combination thereof. \newline \textbf{Document Meta:} type: 10-Q, period\_end\_date: 2022-06-30, fiscal\_year: 2022, period\_focus: Q2. \newline \textbf{Preceding Sentence:} None \newline \textbf{Following Sentence:} If any portion is paid in Canopy Growth Shares, the number of Canopy Growth Shares to be exchanged for each Floating Share shall be determined on the basis of a 30 day volume-weighted average calculation using the Floating Shares (the “Floating Ratio”). \\
$g_{\mathrm{tag}}$ & 1.0000 \\
Label & \path{us-gaap:ShareBasedCompensationArrangementByShareBasedPaymentAwardEquityInstrumentsOtherThanOptionsGrantsInPeriodWeightedAverageGrantDateFairValue} \\
Pred. & \path{custom} \\
CPA Label & \path{custom} or \path{us-gaap:BusinessAcquisitionSharePrice} \\
Reason & \$6.41 is the minimum price at which Canopy Growth may acquire the Floating Shares, not the grant-date fair value of RSUs. \\
\bottomrule
\end{tabular}
\caption{High-gate Tag instance 1.}
\label{tab:audit-tag-high-1}
\end{table*}

\begin{table*}[p]
\centering
\small
\setlength{\tabcolsep}{2pt}
\begin{tabular}{@{} >{\bfseries}p{0.14\textwidth} >{\raggedright\arraybackslash}p{0.82\textwidth} @{}}
\toprule
Field & Value \\
\midrule
Context & \textbf{Target Sentence:} During the three months ended September 30, 2023 and 2022, the Company recognized a credit of \$2.1 million and expense of \$[4.0] million, respectively, of stock compensation expense in connection with PSU awards. \newline \textbf{Document Meta:} type: 10-Q, period\_end\_date: 2023-09-30, fiscal\_year: 2023, period\_focus: Q3. \newline \textbf{Preceding Sentence:} None \newline \textbf{Following Sentence:} During the nine months ended September 30, 2023 and 2022, the Company recognized \$4.3 million and \$6.2 million, respectively, of stock compensation expense in connection with PSU awards. \\
$g_{\mathrm{tag}}$ & 1.0000 \\
Label & \path{custom} \\
Pred. & \path{us-gaap:AllocatedShareBasedCompensationExpense} \\
CPA Label & \path{us-gaap:AllocatedShareBasedCompensationExpense} \\
Reason & The PSU expense is already identified by the award-type axis, so the standard element suffices and the custom extension is unnecessary. \\
\bottomrule
\end{tabular}
\caption{High-gate Tag instance 2.}
\label{tab:audit-tag-high-2}
\end{table*}

\begin{table*}[p]
\centering
\small
\setlength{\tabcolsep}{2pt}
\begin{tabular}{@{} >{\bfseries}p{0.14\textwidth} >{\raggedright\arraybackslash}p{0.82\textwidth} @{}}
\toprule
Field & Value \\
\midrule
Context & \textbf{Target Sentence:} Effective for tax years beginning after December 31, 2022, the IRA imposes a 15\% corporate minimum tax, a [1]\% excise tax on share repurchases, and creates and extends certain tax-related energy incentives. \newline \textbf{Document Meta:} type: 10-K, period\_end\_date: 2023-12-31, fiscal\_year: 2023, period\_focus: FY. \newline \textbf{Preceding Sentence:} On August 16, 2022, the Inflation Reduction Act (``IRA'') was enacted into US law. \newline \textbf{Following Sentence:} Management does not expect the tax-related provisions of the IRA to have a material impact on the Company's consolidated financial statements. \\
$g_{\mathrm{tag}}$ & 1.0000 \\
Label & \path{us-gaap:EffectiveIncomeTaxRateContinuingOperations} \\
Pred. & \path{custom} \\
CPA Label & \path{custom} \\
Reason & 1\% is the excise tax rate on share repurchases under the IRA, not an effective income tax rate. \\
\bottomrule
\end{tabular}
\caption{High-gate Tag instance 3.}
\label{tab:audit-tag-high-3}
\end{table*}

\begin{table*}[p]
\centering
\small
\setlength{\tabcolsep}{2pt}
\begin{tabular}{@{} >{\bfseries}p{0.14\textwidth} >{\raggedright\arraybackslash}p{0.82\textwidth} @{}}
\toprule
Field & Value \\
\midrule
Context & \textbf{Target Sentence:} The Credit Agreements also include several financial covenants, including that (i) IBM will not permit the consolidated net interest expense ratio, for any period of four consecutive fiscal quarters taken as a single accounting period, to be less than 2.20 to 1.0; (ii) the company will not permit its tangible net worth to be less than \$50 million as of the end of the fiscal year and (iii) the company’s leverage ratio cannot be greater than [11] to 1 as of the last day of the fiscal quarter. \newline \textbf{Document Meta:} type: 10-Q, period\_end\_date: 2020-06-30, fiscal\_year: 2020, period\_focus: Q2. \newline \textbf{Preceding Sentence:} The company’s Credit Agreements each contain significant debt covenants, which obligate the company to promptly pay principal and interest, limit the aggregate amount of secured indebtedness and sale and leaseback transactions to 10 percent of IBM’s consolidated net tangible assets, and restrict the ability of the company or IBM to merge or consolidate with a third party, unless certain conditions are met. \newline \textbf{Following Sentence:} The Credit Agreements each contain a cross default provision with respect to other defaulted indebtedness of at least \$500 million. \\
$g_{\mathrm{tag}}$ & 0.0010 \\
Label & \path{custom} \\
Pred. & \path{custom} \\
CPA Label & \path{custom} \\
Reason & The value is the maximum leverage ratio in a debt covenant, for which a custom element is appropriate. \\
\bottomrule
\end{tabular}
\caption{Low-gate Tag instance 1.}
\label{tab:audit-tag-low-1}
\end{table*}

\begin{table*}[p]
\centering
\small
\setlength{\tabcolsep}{2pt}
\begin{tabular}{@{} >{\bfseries}p{0.14\textwidth} >{\raggedright\arraybackslash}p{0.82\textwidth} @{}}
\toprule
Field & Value \\
\midrule
Context & \textbf{Target Sentence:} Prior to the Credit Agreement, (1) borrowings under the Credit Facility accrued interest on a per annum basis at a rate equal to the applicable LIBOR rate plus 2.50\% with no LIBOR floor, and (2) the total borrowing capacity was \$340 million with commitments from a diversified group of [eleven] lenders. \newline \textbf{Document Meta:} type: 10-K, period\_end\_date: 2023-03-31, fiscal\_year: 2023, period\_focus: FY. \newline \textbf{Preceding Sentence:} On August 9, 2021, CSWC entered into the Second Amended and Restated Senior Secured Revolving Credit Agreement (as amended or otherwise modified from time to time, the ``Credit Agreement''). \newline \textbf{Following Sentence:} The Credit Agreement (1) decreased the total borrowing capacity under the Credit Facility to \$335 million with commitments from a diversified group of ten lenders, (2) reduced the interest rate on borrowings to LIBOR plus 2.15\% with no LIBOR floor and removed conditions related thereto as previously set forth in the Amended and Restated Senior Secured Revolving Credit Agreement, and (3) extended the end of the Credit Facility's revolver period from December 21, 2022 to August 9, 2025 and extended the final maturity from December 21, 2023 to August 9, 2026. \\
$g_{\mathrm{tag}}$ & 0.0010 \\
Label & \path{custom} \\
Pred. & \path{custom} \\
CPA Label & \path{custom} \\
Reason & The value is the number of lenders, for which a custom element is appropriate. \\
\bottomrule
\end{tabular}
\caption{Low-gate Tag instance 2.}
\label{tab:audit-tag-low-2}
\end{table*}

\begin{table*}[p]
\centering
\small
\setlength{\tabcolsep}{2pt}
\begin{tabular}{@{} >{\bfseries}p{0.14\textwidth} >{\raggedright\arraybackslash}p{0.82\textwidth} @{}}
\toprule
Field & Value \\
\midrule
Context & \textbf{Target Sentence:} The market condition is achieved in [four] tranches, where 25\% of the options will vest when the trading price of the Company’s stock is above various thresholds of price per share. \newline \textbf{Document Meta:} type: 10-K, period\_end\_date: 2022-12-31, fiscal\_year: 2022, period\_focus: FY. \newline \textbf{Preceding Sentence:} In connection with the Business Combination Agreement with DYNS, on December 19, 2021, Legacy Senti approved 605,451 market awards to its co-founder and Chief Executive Officer, Dr. Timothy Lu, that vest contingent upon the satisfaction of all three of the following conditions: a service condition, a performance condition tied to the consummation of the Merger, and market conditions. \newline \textbf{Following Sentence:} The award and the associated recognition of stock-based compensation expense are contingent on the Merger being consummated. \\
$g_{\mathrm{tag}}$ & 0.0010 \\
Label & \path{custom} \\
Pred. & \path{custom} \\
CPA Label & \path{custom} \\
Reason & The value is the number of vesting tranches, for which a custom element is appropriate. \\
\bottomrule
\end{tabular}
\caption{Low-gate Tag instance 3.}
\label{tab:audit-tag-low-3}
\end{table*}

\begin{table*}[p]
\centering
\small
\setlength{\tabcolsep}{2pt}
\begin{tabular}{@{} >{\bfseries}p{0.14\textwidth} >{\raggedright\arraybackslash}p{0.82\textwidth} @{}}
\toprule
Field & Value \\
\midrule
Context & \textbf{Target Sentence:} The 2022 Plan allows for [32,000,000] shares of Class A Common Stock, plus any shares subject to 2019 Plan awards that expire, are forfeited, canceled or terminated after the adoption of the 2019 Plan. \newline \textbf{Document Meta:} type: 10-K, period\_end\_date: 2023-01-28, fiscal\_year: 2022, period\_focus: FY. \newline \textbf{Preceding Sentence:} Awards under the 2022 Plan may take the form of stock options, stock appreciation rights, restricted stock awards, restricted stock units, performance awards and other share-based awards, or any combination of the foregoing. \newline \textbf{Following Sentence:} No awards were granted under the 2019 or 2011 Plan after the adoption of the 2022 Plan. \\
Tag & \path{us-gaap:ShareBasedCompensationArrangementByShareBasedPaymentAwardNumberOfSharesAvailableForGrant} \\
$g_{\mathrm{time}}$ & 1.0000 \\
Label & instant; past \\
Pred. & instant; current \\
CPA Label & instant; current \\
Reason & The plan was adopted in the past but is currently effective, and the value represents the shares currently available for grant, so instant; current is appropriate. \\
\bottomrule
\end{tabular}
\caption{High-gate Time instance 1.}
\label{tab:audit-time-high-1}
\end{table*}

\begin{table*}[p]
\centering
\small
\setlength{\tabcolsep}{2pt}
\begin{tabular}{@{} >{\bfseries}p{0.14\textwidth} >{\raggedright\arraybackslash}p{0.82\textwidth} @{}}
\toprule
Field & Value \\
\midrule
Context & \textbf{Target Sentence:} The Company allocated the transaction price to each of the separate performance obligations using the relative standalone selling price with \$[58.5] million allocated to the license to the plamotamab program and \$37.6 million allocated to the research services. \newline \textbf{Document Meta:} type: 10-K, period\_end\_date: 2023-12-31, fiscal\_year: 2023, period\_focus: FY. \newline \textbf{Preceding Sentence:} None \newline \textbf{Following Sentence:} None \\
Tag & \path{us-gaap:ContractWithCustomerLiabilityRevenueRecognized} \\
$g_{\mathrm{time}}$ & 1.0000 \\
Label & period; past \\
Pred. & instant; past \\
CPA Label & period; past \\
Reason & The value is the amount of past contract liabilities recognized as revenue, so period; past is appropriate. \\
\bottomrule
\end{tabular}
\caption{High-gate Time instance 2.}
\label{tab:audit-time-high-2}
\end{table*}

\begin{table*}[p]
\centering
\small
\setlength{\tabcolsep}{2pt}
\begin{tabular}{@{} >{\bfseries}p{0.14\textwidth} >{\raggedright\arraybackslash}p{0.82\textwidth} @{}}
\toprule
Field & Value \\
\midrule
Context & \textbf{Target Sentence:} During the first quarter of 2021, the Company received a cash payment of \$[7.5] million related to the first earnout and is reflected within investing activities of discontinued operations on the accompanying consolidated statements of cash flows. \newline \textbf{Document Meta:} type: 10-K, period\_end\_date: 2023-12-31, fiscal\_year: 2023, period\_focus: FY. \newline \textbf{Preceding Sentence:} The Divestiture resulted in a pre-tax gain of \$92.5 million for the year ended December 31, 2020. \newline \textbf{Following Sentence:} During the third quarter of 2021, the Company finalized its net working capital adjustments associated with the Divestiture. \\
Tag & \path{us-gaap:DisposalGroupIncludingDiscontinuedOperationConsideration} \\
$g_{\mathrm{time}}$ & 1.0000 \\
Label & instant; current \\
Pred. & period; past \\
CPA Label & period; past \\
Reason & The earnout cash was received in the first quarter of 2021, so it is a value for a past period rather than a point-in-time value at the current period end. \\
\bottomrule
\end{tabular}
\caption{High-gate Time instance 3.}
\label{tab:audit-time-high-3}
\end{table*}

\begin{table*}[p]
\centering
\small
\setlength{\tabcolsep}{2pt}
\begin{tabular}{@{} >{\bfseries}p{0.14\textwidth} >{\raggedright\arraybackslash}p{0.82\textwidth} @{}}
\toprule
Field & Value \\
\midrule
Context & \textbf{Target Sentence:} At June 30, 2023 and December 31, 2022, assets of the VIEs totaled \$472,108 and \$497,921, and liabilities of the VIEs totaled \$[465,860] and \$453,952, respectively. \newline \textbf{Document Meta:} type: 10-Q, period\_end\_date: 2023-06-30, fiscal\_year: 2023, period\_focus: Q2. \newline \textbf{Preceding Sentence:} (1)The condensed consolidated balance sheets include assets and liabilities of consolidated variable interest entities, or VIEs. \newline \textbf{Following Sentence:} See Note 3 - Variable Interest Entities for additional information. \\
Tag & \path{us-gaap:Liabilities} \\
$g_{\mathrm{time}}$ & 0.0004 \\
Label & instant; current \\
Pred. & instant; current \\
CPA Label & instant; current \\
Reason & The date in the sentence is the end of the current reporting period. \\
\bottomrule
\end{tabular}
\caption{Low-gate Time instance 1.}
\label{tab:audit-time-low-1}
\end{table*}

\begin{table*}[p]
\centering
\small
\setlength{\tabcolsep}{2pt}
\begin{tabular}{@{} >{\bfseries}p{0.14\textwidth} >{\raggedright\arraybackslash}p{0.82\textwidth} @{}}
\toprule
Field & Value \\
\midrule
Context & \textbf{Target Sentence:} At September 30, 2023 and December 31, 2022, assets of the VIEs totaled \$462,004 and \$497,921, and liabilities of the VIEs totaled \$[453,629] and \$453,952, respectively. \newline \textbf{Document Meta:} type: 10-Q, period\_end\_date: 2023-09-30, fiscal\_year: 2023, period\_focus: Q3. \newline \textbf{Preceding Sentence:} (1)The condensed consolidated balance sheets include assets and liabilities of consolidated variable interest entities, or VIEs. \newline \textbf{Following Sentence:} See Note 4 - Variable Interest Entities for additional information. \\
Tag & \path{us-gaap:Liabilities} \\
$g_{\mathrm{time}}$ & 0.0004 \\
Label & instant; current \\
Pred. & instant; current \\
CPA Label & instant; current \\
Reason & The date in the sentence is the end of the current reporting period. \\
\bottomrule
\end{tabular}
\caption{Low-gate Time instance 2.}
\label{tab:audit-time-low-2}
\end{table*}

\begin{table*}[p]
\centering
\small
\setlength{\tabcolsep}{2pt}
\begin{tabular}{@{} >{\bfseries}p{0.14\textwidth} >{\raggedright\arraybackslash}p{0.82\textwidth} @{}}
\toprule
Field & Value \\
\midrule
Context & \textbf{Target Sentence:} Of the \$[1.71] billion of securities in an unrealized loss position, \$383.1 million were in an unrealized loss position for 12 months or longer. \newline \textbf{Document Meta:} type: 10-Q, period\_end\_date: 2022-09-30, fiscal\_year: 2022, period\_focus: Q3. \newline \textbf{Preceding Sentence:} At September 30, 2022, Park’s debt securities portfolio consisted of \$1.74 billion of securities, \$1.71 billion of which were in an unrealized loss position with unrealized losses of \$151.4 million. \newline \textbf{Following Sentence:} Of the \$151.4 million in unrealized losses, an aggregate of \$93.5 million were related to Park's ``Obligations of U.S. Treasury and other U.S. Government sponsored entities'' portfolio and Park's ``U.S. Government sponsored entities' asset-backed securities'' portfolio. \\
Tag & \path{us-gaap:DebtSecuritiesAvailableForSaleUnrealizedLossPosition} \\
$g_{\mathrm{time}}$ & 0.0004 \\
Label & instant; current \\
Pred. & instant; current \\
CPA Label & instant; current \\
Reason & The date in the sentence is the end of the current reporting period. \\
\bottomrule
\end{tabular}
\caption{Low-gate Time instance 3.}
\label{tab:audit-time-low-3}
\end{table*}

\begin{table*}[p]
\centering
\small
\setlength{\tabcolsep}{2pt}
\begin{tabular}{@{} >{\bfseries}p{0.14\textwidth} >{\raggedright\arraybackslash}p{0.82\textwidth} @{}}
\toprule
Field & Value \\
\midrule
Context & \textbf{Target Sentence:} The risk-free rates utilized were [2.454]\% and 1.123\% for the valuations as of March 31, 2022 and December 31, 2021, respectively. \newline \textbf{Document Meta:} type: 10-Q, period\_end\_date: 2022-03-31, fiscal\_year: 2022, period\_focus: Q1. \newline \textbf{Preceding Sentence:} This assumption considers observed historical stock price volatility of other companies operating in the same or similar industry as the Company over a period similar to the remaining term of the Private Warrants, as well as the volatility implied by the traded options of the Company. \newline \textbf{Following Sentence:} None \\
Tag & \path{us-gaap:DerivativeLiabilityMeasurementInput} \\
$g_{\mathrm{scale}}$ & 1.0000 \\
Label & 0 \\
Pred. & $-2$ \\
CPA Label & $-2$ \\
Reason & The risk-free rate should be recorded as 0.02454 but is recorded as 2.454, so the scale is also wrong. \\
\bottomrule
\end{tabular}
\caption{High-gate Scale instance 1.}
\label{tab:audit-scale-high-1}
\end{table*}

\begin{table*}[p]
\centering
\small
\setlength{\tabcolsep}{2pt}
\begin{tabular}{@{} >{\bfseries}p{0.14\textwidth} >{\raggedright\arraybackslash}p{0.82\textwidth} @{}}
\toprule
Field & Value \\
\midrule
Context & \textbf{Target Sentence:} The risk-free rates utilized were 2.454\% and [1.123]\% for the valuations as of March 31, 2022 and December 31, 2021, respectively. \newline \textbf{Document Meta:} type: 10-Q, period\_end\_date: 2022-03-31, fiscal\_year: 2022, period\_focus: Q1. \newline \textbf{Preceding Sentence:} This assumption considers observed historical stock price volatility of other companies operating in the same or similar industry as the Company over a period similar to the remaining term of the Private Warrants, as well as the volatility implied by the traded options of the Company. \newline \textbf{Following Sentence:} None \\
Tag & \path{us-gaap:DerivativeLiabilityMeasurementInput} \\
$g_{\mathrm{scale}}$ & 1.0000 \\
Label & 0 \\
Pred. & $-2$ \\
CPA Label & $-2$ \\
Reason & The risk-free rate should be recorded as 0.01123 but is recorded as 1.123, so the scale is also wrong. \\
\bottomrule
\end{tabular}
\caption{High-gate Scale instance 2.}
\label{tab:audit-scale-high-2}
\end{table*}

\begin{table*}[p]
\centering
\small
\setlength{\tabcolsep}{2pt}
\begin{tabular}{@{} >{\bfseries}p{0.14\textwidth} >{\raggedright\arraybackslash}p{0.82\textwidth} @{}}
\toprule
Field & Value \\
\midrule
Context & \textbf{Target Sentence:} The Company recognized revenue of \$[0.2] million and \$0.8 million during the three and six months ended June 30, 2021, respectively. \newline \textbf{Document Meta:} type: 10-Q, period\_end\_date: 2022-06-30, fiscal\_year: 2022, period\_focus: Q2. \newline \textbf{Preceding Sentence:} During the three and six months ended June 30, 2021, the total costs expected to be incurred to satisfy the performance obligation under the 2017 Roche Agreement decreased by \$0.1 million and \$0.3 million, respectively. \newline \textbf{Following Sentence:} None \\
Tag & \path{us-gaap:ContractWithCustomerLiabilityRevenueRecognized} \\
$g_{\mathrm{scale}}$ & 1.0000 \\
Label & 3 \\
Pred. & 6 \\
CPA Label & 6 \\
Reason & \$0.2 million should be recorded as 200,000 but is recorded as 200. \\
\bottomrule
\end{tabular}
\caption{High-gate Scale instance 3.}
\label{tab:audit-scale-high-3}
\end{table*}

\begin{table*}[p]
\centering
\small
\setlength{\tabcolsep}{2pt}
\begin{tabular}{@{} >{\bfseries}p{0.14\textwidth} >{\raggedright\arraybackslash}p{0.82\textwidth} @{}}
\toprule
Field & Value \\
\midrule
Context & \textbf{Target Sentence:} In Europe, the Nasdaq Nordic and Nasdaq Baltic exchanges, together with Nasdaq First North, were home to 1,152 listed companies with a combined market capitalization of approximately \$[2.4] trillion.Our IR \& ESG Services business includes our Investor Relations Intelligence and Governance Solutions businesses, which serve both public and private companies and organizations. \newline \textbf{Document Meta:} type: 10-Q, period\_end\_date: 2021-06-30, fiscal\_year: 2021, period\_focus: Q2. \newline \textbf{Preceding Sentence:} The combined market capitalization was approximately \$25.8 trillion. \newline \textbf{Following Sentence:} Our public company clients can be companies listed on our exchanges or other U.S. and global exchanges. \\
Tag & \path{custom} \\
$g_{\mathrm{scale}}$ & 0.0000 \\
Label & 12 \\
Pred. & 12 \\
CPA Label & 12 \\
Reason & Trillion corresponds to a scale of 12. \\
\bottomrule
\end{tabular}
\caption{Low-gate Scale instance 1.}
\label{tab:audit-scale-low-1}
\end{table*}

\begin{table*}[p]
\centering
\small
\setlength{\tabcolsep}{2pt}
\begin{tabular}{@{} >{\bfseries}p{0.14\textwidth} >{\raggedright\arraybackslash}p{0.82\textwidth} @{}}
\toprule
Field & Value \\
\midrule
Context & \textbf{Target Sentence:} In Europe, the Nasdaq Nordic and Nasdaq Baltic exchanges, together with Nasdaq First North, were home to 1,253 listed companies with a combined market capitalization of approximately \$[1.5] trillion. \newline \textbf{Document Meta:} type: 10-Q, period\_end\_date: 2022-09-30, fiscal\_year: 2022, period\_focus: Q3. \newline \textbf{Preceding Sentence:} The combined market capitalization was approximately \$19.3 trillion. \newline \textbf{Following Sentence:} None \\
Tag & \path{custom} \\
$g_{\mathrm{scale}}$ & 0.0000 \\
Label & 12 \\
Pred. & 12 \\
CPA Label & 12 \\
Reason & Trillion corresponds to a scale of 12. \\
\bottomrule
\end{tabular}
\caption{Low-gate Scale instance 2.}
\label{tab:audit-scale-low-2}
\end{table*}

\begin{table*}[p]
\centering
\small
\setlength{\tabcolsep}{2pt}
\begin{tabular}{@{} >{\bfseries}p{0.14\textwidth} >{\raggedright\arraybackslash}p{0.82\textwidth} @{}}
\toprule
Field & Value \\
\midrule
Context & \textbf{Target Sentence:} Excludes deposit liabilities with no defined or contractual maturity of \$1.2 trillion and \$[1.3] trillion at June 30, 2023, and December 31, 2022, respectively. \newline \textbf{Document Meta:} type: 10-Q, period\_end\_date: 2023-06-30, fiscal\_year: 2023, period\_focus: Q2. \newline \textbf{Preceding Sentence:} None \newline \textbf{Following Sentence:} None \\
Tag & \path{custom} \\
$g_{\mathrm{scale}}$ & 0.0000 \\
Label & 12 \\
Pred. & 12 \\
CPA Label & 12 \\
Reason & Trillion corresponds to a scale of 12. \\
\bottomrule
\end{tabular}
\caption{Low-gate Scale instance 3.}
\label{tab:audit-scale-low-3}
\end{table*}

\begin{table*}[p]
\centering
\small
\setlength{\tabcolsep}{2pt}
\begin{tabular}{@{} >{\bfseries}p{0.14\textwidth} >{\raggedright\arraybackslash}p{0.82\textwidth} @{}}
\toprule
Field & Value \\
\midrule
Context & \textbf{Target Sentence:} As a result of the transaction, the company recorded a \$[72.1] million loss within the consolidated statements of comprehensive loss for the year ended December 31, 2021. \newline \textbf{Document Meta:} type: 10-K, period\_end\_date: 2022-12-31, fiscal\_year: 2022, period\_focus: FY. \newline \textbf{Preceding Sentence:} Management determined that because the IPR\&D asset did not have technological feasibility, it was immediately expensed. \newline \textbf{Following Sentence:} This loss has a net impact of \$30.2 million to Surgalign, and \$41.9 million impact to INN. \\
Tag & \path{us-gaap:BusinessCombinationAcquisitionRelatedCosts} \\
$g_{\mathrm{sign}}$ & 1.0000 \\
Label & positive \\
Pred. & negative \\
CPA Label & negative, with \path{us-gaap:VariableInterestEntityInitialConsolidationGainOrLoss} \\
Reason & INN is a variable interest entity, so the initial-consolidation gain-or-loss element fits better than the acquisition-cost element. This element has a credit balance, so the loss is negative. With the original tag alone, a positive sign would be consistent. \\
\bottomrule
\end{tabular}
\caption{High-gate Sign instance 1.}
\label{tab:audit-sign-high-1}
\end{table*}

\begin{table*}[p]
\centering
\small
\setlength{\tabcolsep}{2pt}
\begin{tabular}{@{} >{\bfseries}p{0.14\textwidth} >{\raggedright\arraybackslash}p{0.82\textwidth} @{}}
\toprule
Field & Value \\
\midrule
Context & \textbf{Target Sentence:} During the years ended December 31, 2022 and 2021, TEPH unwound interest rate swaps and caps with an aggregate notional amount of \$515.4 million and \$121.3 million, respectively, and recorded a realized gain of \$28.3 million and a realized loss of \$[2.2] million, respectively. \newline \textbf{Document Meta:} type: 10-K, period\_end\_date: 2022-12-31, fiscal\_year: 2022, period\_focus: FY. \newline \textbf{Preceding Sentence:} In October 2023, the notional amount of the interest rate swaps and caps will begin decreasing to match TEPH's estimated quarterly principal payments on the debt. \newline \textbf{Following Sentence:} None \\
Tag & \path{us-gaap:LossOnDerivativeInstrumentsPretax} \\
$g_{\mathrm{sign}}$ & 1.0000 \\
Label & positive \\
Pred. & negative \\
CPA Label & positive \\
Reason & The element denotes a loss, so a positive value is correct. \\
\bottomrule
\end{tabular}
\caption{High-gate Sign instance 2.}
\label{tab:audit-sign-high-2}
\end{table*}

\begin{table*}[p]
\centering
\small
\setlength{\tabcolsep}{2pt}
\begin{tabular}{@{} >{\bfseries}p{0.14\textwidth} >{\raggedright\arraybackslash}p{0.82\textwidth} @{}}
\toprule
Field & Value \\
\midrule
Context & \textbf{Target Sentence:} During the nine months ended September 30, 2022 and 2021, EZOP unwound interest rate swaps and caps with an aggregate notional amount of \$360.2 million and \$131.7 million, respectively, and recorded a realized gain of \$19.6 million and a realized loss of \$[68,000], respectively. \newline \textbf{Document Meta:} type: 10-Q, period\_end\_date: 2022-09-30, fiscal\_year: 2022, period\_focus: Q3. \newline \textbf{Preceding Sentence:} In July 2022, the notional amount of the interest rate swaps and caps began decreasing to match EZOP's estimated monthly principal payments on the debt. \newline \textbf{Following Sentence:} None \\
Tag & \path{us-gaap:LossOnDerivativeInstrumentsPretax} \\
$g_{\mathrm{sign}}$ & 1.0000 \\
Label & positive \\
Pred. & negative \\
CPA Label & positive \\
Reason & The element denotes a loss, so a positive value is correct. \\
\bottomrule
\end{tabular}
\caption{High-gate Sign instance 3.}
\label{tab:audit-sign-high-3}
\end{table*}

\begin{table*}[p]
\centering
\small
\setlength{\tabcolsep}{2pt}
\begin{tabular}{@{} >{\bfseries}p{0.14\textwidth} >{\raggedright\arraybackslash}p{0.82\textwidth} @{}}
\toprule
Field & Value \\
\midrule
Context & \textbf{Target Sentence:} (4)Other financing obligations included \$20.4 million related to insurance premium notes and certain secured borrowings that generally bore interest rates ranging from 2.40\% to [6.40]\%, and \$14.5 million related to outstanding borrowings on the construction loan agreements that bore a weighted average interest rate of 6.778\% during the period ended March 31, 2023. \newline \textbf{Document Meta:} type: 10-Q, period\_end\_date: 2023-03-31, fiscal\_year: 2023, period\_focus: Q1. \newline \textbf{Preceding Sentence:} None \newline \textbf{Following Sentence:} Other financing obligations included \$20.7 million related to insurance premium notes and certain secured borrowings that generally bore interest rates ranging from 2.40\% to 5.84\%, and \$7.4 million related to outstanding borrowings on the construction loan agreements that bore a weighted average interest rate of 5.634\% during the period ended December 31, 2022. \\
Tag & \path{us-gaap:DebtInstrumentInterestRateStatedPercentage} \\
$g_{\mathrm{sign}}$ & 0.0003 \\
Label & positive \\
Pred. & positive \\
CPA Label & positive \\
Reason & The value is an interest rate and is positive. \\
\bottomrule
\end{tabular}
\caption{Low-gate Sign instance 1.}
\label{tab:audit-sign-low-1}
\end{table*}

\begin{table*}[p]
\centering
\small
\setlength{\tabcolsep}{2pt}
\begin{tabular}{@{} >{\bfseries}p{0.14\textwidth} >{\raggedright\arraybackslash}p{0.82\textwidth} @{}}
\toprule
Field & Value \\
\midrule
Context & \textbf{Target Sentence:} (4)Other financing obligations included \$20.4 million related to insurance premium notes and certain secured borrowings that generally bore interest rates ranging from [2.40]\% to 6.40\%, and \$14.5 million related to outstanding borrowings on the construction loan agreements that bore a weighted average interest rate of 6.778\% during the period ended March 31, 2023. \newline \textbf{Document Meta:} type: 10-Q, period\_end\_date: 2023-03-31, fiscal\_year: 2023, period\_focus: Q1. \newline \textbf{Preceding Sentence:} None \newline \textbf{Following Sentence:} Other financing obligations included \$20.7 million related to insurance premium notes and certain secured borrowings that generally bore interest rates ranging from 2.40\% to 5.84\%, and \$7.4 million related to outstanding borrowings on the construction loan agreements that bore a weighted average interest rate of 5.634\% during the period ended December 31, 2022. \\
Tag & \path{us-gaap:DebtInstrumentInterestRateStatedPercentage} \\
$g_{\mathrm{sign}}$ & 0.0003 \\
Label & positive \\
Pred. & positive \\
CPA Label & positive \\
Reason & The value is an interest rate and is positive. \\
\bottomrule
\end{tabular}
\caption{Low-gate Sign instance 2.}
\label{tab:audit-sign-low-2}
\end{table*}

\begin{table*}[p]
\centering
\small
\setlength{\tabcolsep}{2pt}
\begin{tabular}{@{} >{\bfseries}p{0.14\textwidth} >{\raggedright\arraybackslash}p{0.82\textwidth} @{}}
\toprule
Field & Value \\
\midrule
Context & \textbf{Target Sentence:} On June 2, 2020, the Company issued \$747.5 million aggregate principal amount of [0.125]\% convertible senior notes due 2025 (the “2025 Notes”) in a private placement to qualified institutional buyers pursuant to Rule 144A under the Securities Act of 1933, as amended (“Securities Act”). \newline \textbf{Document Meta:} type: 10-Q, period\_end\_date: 2023-06-30, fiscal\_year: 2023, period\_focus: Q2. \newline \textbf{Preceding Sentence:} None \newline \textbf{Following Sentence:} The total net proceeds from the sale of the 2025 Notes, after deducting the initial purchasers’ discounts and debt issuance costs, were approximately \$730.2 million. \\
Tag & \path{us-gaap:DebtInstrumentInterestRateStatedPercentage} \\
$g_{\mathrm{sign}}$ & 0.0003 \\
Label & positive \\
Pred. & positive \\
CPA Label & positive \\
Reason & The value is an interest rate and is positive. \\
\bottomrule
\end{tabular}
\caption{Low-gate Sign instance 3.}
\label{tab:audit-sign-low-3}
\end{table*}

\end{document}